\documentclass[10pt]{article} 
\usepackage[accepted]{tmlr}

\usepackage{amsmath,amsfonts,bm}

\def\eqref#1{equation~\ref{#1}}

\def\1{\bm{1}}

\def\eps{{\epsilon}}

\DeclareMathAlphabet{\mathsfit}{\encodingdefault}{\sfdefault}{m}{sl}
\SetMathAlphabet{\mathsfit}{bold}{\encodingdefault}{\sfdefault}{bx}{n}

\newcommand{\E}{\mathbb{E}}

\newcommand{\R}{\mathbb{R}}

\usepackage{microtype}
\usepackage{graphicx}
\usepackage{caption}
\usepackage{subcaption}
\usepackage{wrapfig}
\usepackage[font=small,labelfont=bf]{caption}

\usepackage{xcolor}
\usepackage{hyperref}
\usepackage{url}            
\usepackage{booktabs}       
\usepackage{amsfonts}       
\usepackage{nicefrac}       
\usepackage{microtype}      
\usepackage{xcolor}         
\usepackage{amsmath}
\usepackage{amsthm}
\usepackage{amssymb}
\usepackage{multirow}
\usepackage{enumitem}
\usepackage{array} 
\usepackage{wrapfig}
\usepackage{algorithm}
\usepackage{algorithmic}
\usepackage{xspace}
\usepackage{float}
\usepackage{mymacros}
\usepackage{cleveref}
\usepackage[textsize=small]{todonotes}

\usepackage{hyperref}
\usepackage{url}
\newcommand{\coe}{\texttt{ABC}\xspace}
\newcommand{\updated}[1]{{#1}}

\newenvironment{Updated}{\par}{\par}

\usepackage{xcolor}
\definecolor{ppurple}{HTML}{603a70}
\usepackage[most,skins,theorems]{tcolorbox}
\tcbset{
  aibox/.style={
    width=\linewidth,
    top=8pt,
    bottom=4pt,
    colback=ppurple!6!white,
    colframe=ppurple,
    colbacktitle=ppurple,
    enhanced,
    center,
    attach boxed title to top left={yshift=-0.1in,xshift=0.15in},
    boxed title style={boxrule=0pt,colframe=white,},
  }
}
\newtcolorbox{AIbox}[2][]{aibox,title=#2,#1}

\title{Agreement-Based Cascading for Efficient Inference}

\author{\name Steven Kolawole\thanks{Equal contribution.}\footnotemark[1] \email skolawol@andrew.cmu.edu \\
      \addr Carnegie Mellon University
      \AND
      \name Don Dennis\footnotemark[1] \email dondennis@cmu.edu \\
      \addr Carnegie Mellon University
      \AND
      \name Ameet Talwalkar \email talwalkar@cmu.edu \\
      \addr Carnegie Mellon University
      \AND
      \name Virginia Smith \email smithv@cmu.edu \\
      \addr Carnegie Mellon University
    }

\def\month{07}  
\def\year{2025} 
\def\openreview{\url{https://openreview.net/forum?id=jn9B7LMlzk}}

\begin{document}

\maketitle

\vspace{-.1in}
\begin{abstract}
Adaptive inference schemes reduce the cost of machine
learning inference by assigning smaller models to easier examples, attempting to
avoid invocation of larger models when possible. In this work we explore a simple, effective adaptive inference technique we term
    \textit{Agreement-Based Cascading (\coe)}. \coe builds a cascade of models of increasing size/complexity and uses agreement between ensembles of models at each level of the cascade as a basis for data-dependent routing. 
    Although ensemble execution introduces additional expense, we show that these costs can be easily offset in practice due to large expected differences in model sizes, parallel inference execution capabilities, and accuracy benefits of ensembling. We examine \coe theoretically and empirically in terms of these parameters, showing that the approach can reliably act as a drop-in replacement for existing models and surpass the best single model it aims to replace in terms of both efficiency and accuracy. Additionally, we explore the performance of \coe relative to existing cascading methods in three common scenarios: (1) edge-to-cloud inference, where \coe~reduces
    communication costs by up to $14\times$; (2) cloud-based model serving, where it achieves a $3\times$ reduction in rental costs; and (3)
    inference via model API services, where \coe achieves a $2$-$25\times$ reduction in average price per token/request relative to state-of-the-art LLM cascades.
\end{abstract}

\section{Introduction}

The high cost of inference associated with deploying large machine
learning (ML) models presents a significant barrier to their adoption \citep{strubell2020energy, Kaplan2020ScalingLF}.
As models continue to increase in size, practitioners are often faced with
investing substantial resources in updating existing deployments or settling for
lower-performing alternatives.
However, for many applications, it has been shown that a considerable portion of the data seen
during inference can be effectively evaluated using small
models rather than large, state-of-the-art models~\citep{chen_frugalml_2020, jitkrittum_when_2023}. This means
that if we can identify the subset of data samples that can be accurately evaluated by more inexpensive models, average inference costs can be reduced.
This problem is often referred to as \emph{adaptive inference}, where the cost of inference adapts to some notion
of `difficulty' of each example seen at inference time. 



A natural approach for {adaptive inference} is to \textit{cascade} over a
set of potential models, starting from the least expensive and moving to more
expensive models based on some \textit{deferral rule}~\citep{ rowley_neural_1998,
viola_rapid_2001, soo_object_2014}. A common cascade construction is to use a
Pareto-efficient set of models and an easy-to-compute deferral rule such as the
confidence scores of the models predictions~\citep{viola_robust_2004,
wang_idk_2018, wang2021wisdom}.  
Recently, cascading has seen renewed interest in the ML community with the advent of large language models (LLMs) and vision foundation models
\citep[e.g.,][]{chen2023frugalgpt, gupta2024language}.
%
Recent approaches consider a variety of techniques to construct model
cascades,
including designing novel model
architectures
that have built-in cascading capabilities
\citep{cai_once-for-all_2020, devvrit_matformer_2023, khare_superserve_2023}, or
learning routing schemes/deferral rules that require
data-dependent training for every task considered \citep{chen2023frugalgpt, dinghybrid}. 
While these approaches can improve accuracy/efficiency trade-offs, they may also introduce significant computational overhead in setup and training/finetuning costs.



\begin{figure*}[t]
    \vspace{-3.5em}
  \centering
    \begin{subfigure}{0.55\textwidth}
      \includegraphics[clip,width=\linewidth]{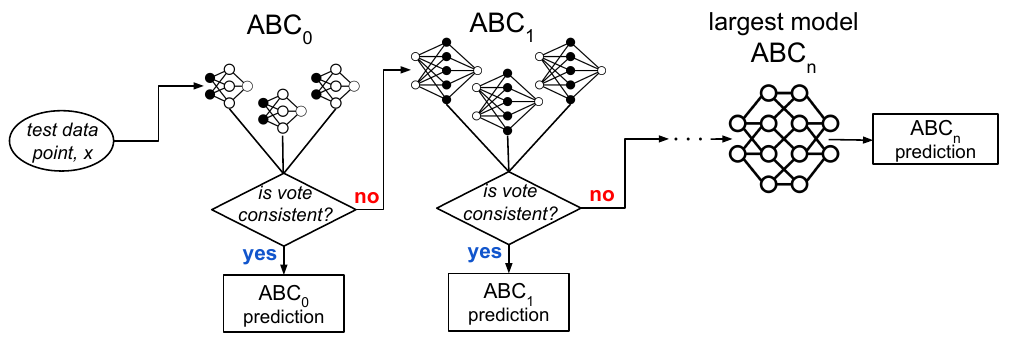}
      \subcaption{}
  \end{subfigure}
  \begin{subfigure}{0.44\textwidth}
     \centering
         \includegraphics[width=\linewidth]{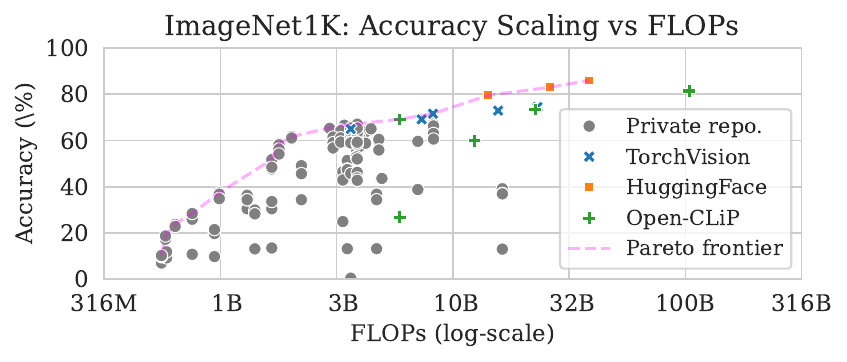}
      \subcaption{}
  \end{subfigure}
  
  \vspace{-.1in}

  \caption{\textbf{(a)} Agreement-Based Cascading (\coe): \coe
  introduces a data-dependent routing scheme that uses agreement amongst an
  ensemble of models to determine whether to cascade to larger models. If the
  predictions of smaller ensembles do not align, the cascade moves to the next
  tier of larger models, continuing until agreement is reached or the largest
  model(s) are used. This can reduce cost
  by limiting the use of the largest models to cases where smaller models cannot reach consensus.
   \textbf{(b)} \coe is a natural baseline for adaptive inference due to  (i) the vast number of pretrained models 
  available to ML practitioners today; (ii) the fact that even small accuracy gains often require
  an order-of-magnitude increase in FLOPs, mirroring proposed scaling laws and resulting in large differences in model sizes between cascade tiers~\citep{hestness2017deep,
  henighan2020scaling, madaan_automix_2023}. The pink dashed line represents the
  Pareto-optimal frontier, showing the models with the highest accuracy for a
  given computational budget. 
  We show that \coe can effectively improve this frontier---allowing practitioners
  to achieve high accuracy without incurring the full computational cost of the
  largest models by invoking smaller models for `easier' samples.
  }

  \label{fig:method_diagram} 
  \label{fig:pareto}
  \vspace{-3mm}
\end{figure*}




In this work, we instead investigate a simple, \textit{training-free} cascade scheme that uses the
{agreement} among an {ensemble} of existing models at each cascade
level as its deferral rule. 
We refer to this approach as \textit{Agreement-Based Cascading (\coe)}. Intuitively, when the outputs of the ensemble at a particular cascade-level do not align, \coe triggers cascading, and inference is attempted at the next tier of (larger) models (see Figure~\ref{fig:method_diagram}). 
This scheme has been explored historically for specialized applications such as face
detection~\citep{10.1007/11558484_4, Zuo_cascaded, Susnjak2012AdaptiveCO},
where ensembles of binary face detectors focus on subregions of the
face and combine their output using decision networks. However, to the
best of our knowledge, our work is the first to study the use of ensemble
agreement as a deferral mechanism for more general, modern machine
learning workloads.


\begin{Updated}
We point to a few trends in ML that make \coe a particularly attractive approach for model cascading. In particular, while using an ensemble of models at each level may initially appear to increase overall inference costs, it is natural to believe this simple baseline could excel in real-world applications due to:
(1) the growing ease of obtaining pretrained models of various sizes and accuracies
due to the rise of model registries \citep{wolf2019huggingface} as well as flexible
compression schemes \citep{zhu2023survey,dennis_progressive_2023}; (2) scaling laws that predict a large increase in inference
cost for every 
marginal increase in accuracy
~\citep{hestness2017deep} (Figure~\ref{fig:pareto}); and (3) an increased ability to execute ensembles of
relatively small models in parallel, with minimal additional
costs \citep{fern2003online, Kim2023AnLC, Miao2023TowardsEG} (Figure~\ref{fig:syntheticparallel}).


With these motivations in mind, we 
rigorously study
\coe as a baseline for adaptive inference in modern ML workloads. Overall, we make the following contributions:
\begin{itemize}[itemsep=0pt, topsep=0pt, partopsep=0pt, leftmargin=*]
    \item  We propose \coe as a training-free, deferral-based cascading approach that leverages agreement among ensembles of existing models, removing the need for additional routing networks or specialized architectures.
    
    

    \item We theoretically characterize cases where \coe can replace  an
    existing model deployment without affecting the accuracy, and define safe
    deferral rules as sufficiency conditions for their existence. We further
    characterize the expected accuracy and inference costs
    in this setting.
    
    \item We empirically evaluate \coe on a wide range of image and language tasks and find that \coe not only improves efficiency, but also accuracy, compared to the model that it aims to replace. We
    then consider the performance of \coe relative to existing cascading methods in common inference scenarios, including (1) edge-to-cloud inference where
    \coe reduces communication costs by up to $14\times$, (2) model-serving on heterogeneous GPUs, where \coe reduces rental costs by
    up to $3\times$ and (3) inference using black-box access to model API
    services, where \coe shows up to a $25\times$ reduction in average price
    per token.
\end{itemize}
\end{Updated}

\section{Related Work}
\label{sec:relatedwork}

Adaptive inference schemes have been a topic of interest in machine learning for many years. This section discusses three main approaches to cascading and adaptive inference: score-based methods, trained routers, and dynamic networks. We highlight how \coe's approach relates to and differs from these existing methods.


\subsection{Cascading using Score-based Deferrals} 
Traditional cascading methods often rely on simple \textit{score-based metrics} for deferral decisions.
We compare to the recent Wisdom-of-Committees method of~\citet{wang2021wisdom} as a general, representative method in this category but note that specialized instantiations of this approach
 have been applied across various domains in prior work, for instance, in object detection
\citep{rowley_neural_1998, viola_robust_2004, wang_cascade_2011, cai_learning_2015, angelova_real-time_2015, streeter_approximation_2018}, image classification \citep{wang_idk_2018, wang2021wisdom}, and text classification \citep{li_cascadebert_2021, mamou_tangobert_2022, varshney_model_2022, lebovitz_efficient_2023}. 
In these settings, deferrals within cascading systems use metrics such as confidence scores, entropy, and probabilities \citep{gangrade_selective_2021, geifman_selectivenet_2019, narasimhan_post-hoc_2022}.  
While using confidence scores of existing models is inexpensive, these scores are required to be well-calibrated, which is less common for off-the-shelf models \citep{pmlr-v70-guo17a, enomoro2021learning}. Recent methods have thus considered a variety of techniques to enhance score-based deferrals, such as explicitly adding deferrals as a prediction~\citep{wang_idk_2018}, difficulty-aware regularization~\citep{li_cascadebert_2021}, temperature scaling~\citep{wang2023tabi}, and calibration mechanisms~\citep{nie2024online}.

Recently, \citet{jitkrittum_when_2023} explored the specific practical settings under which confidence-based deferral in cascades could suffer; these settings include: scenarios where downstream models are specialists (the error probability of the downstream model is highly non-uniform across samples), when samples are subject to label noise, and in the presence of a distribution shift between the train and test set. Notably, in all these failure modes, where confidence-based cascades tend to be inadequate, our approach is likely to offer improvements, as ensembles are known to help induce diversity, enable robustness to noise, and mitigate issues of distribution shift \citep{gontijo-lopes_no_2022, doi:10.1080/095400996116785, dietterich2000ensemble, dvzeroski2004combining}.

\subsection{Cascades with Routing Procedures}

Other methods avoid confidence scores entirely by
training procedures independent of the model set to route instances as needed.
This trend has gained prominence with the rise of black-box model API services for LLMs, where prediction scores are often unavailable.
\citet{guan_energy-efficient_2018} developed a selection module trained to
determine the best-fit classifiers for data instances. 
\citet{yue2024large} (MoT LLM Cascade) used sampling and consistency checking to determine when to defer to a  high-cost model. 
AutoMix \citep{madaan_automix_2023} proposed a few-shot self-verification mechanism, similar to \citet{yue2024large}, but also introduced a Markov-based meta-verifier
for cascading
in context-grounded tasks. 
FrugalGPT
\citep{chen2023frugalgpt} leveraged a
cascade
strategy that triages incoming queries using a 
DistilBERT router and scoring function. HybridLLM \citep{dinghybrid} used a fine-tuned DeBERTa \citep{he2020deberta} to route queries to models based on the predicted query difficulty and the desired quality level. On the same note, OrchestraLLM~\citep{lee2023orchestrallm}---using hand-labeled data to create expert model pools---selects
model to query based on embedding distances between the pools' instances and the test instances. 
Other methods also used some form of trained router, including RouteLLM~\citep{ong2024routellm}, `Fly-swat or cannon'~\citep{vsakota2024fly}, and \citet{shnitzer_large_2023}.

As we further discuss in Section~\ref{sec:api_settings} (where we empirically compare to FrugalGPT, AutoMix, and MoT LLM Cascade), these more sophisticated methods 
involve complex setups,
data-dependent training, and increased computational overhead. 
They often
require retraining routers for each new task, dataset, or model,
limiting adaptability
to unseen data distributions or new models without
incurring further costs. 
In 
contrast, \coe provides a
simpler, flexible, widely applicable alternative
at no additional training or setup cost. 

\subsection{Dynamic and Adaptive Networks} 
Routing procedures typically train an
independent routing mechanism, keeping the Pareto-set of models untouched.
However, for many medium-scale applications, methods have been explored that
learn a Pareto-set of models and stopping criterion together. For example, early exit
methods of \citet{bolukbasi2017adaptive, huang_multi-scale_2018, shafiee2018efficient, wang_skipnet_2018, hutriple,
xin2020deebert, geng2021romebert, zhou2020bert, liu2020fastbert, schuster_confident_2022}, subnetworks extraction methods~\citep{yu_slimmable_2018, yu_universally_2019, Chen_2021_ICCV,
hou_dynabert_2020, han_dynamic_2022, devvrit_matformer_2023}, composition-based
methods \citep{NEURIPS2020_arun, dennis_progressive_2023, du_compositional_2024} are some of the more recent examples. While these methods achieve impressive efficiency gains, they require training specialized architectures from scratch with adaptive capabilities.

Our approach diverges from these methods, as it does not involve altering model architectures or retraining from scratch. Instead, \coe~leverages existing models as they are, utilizing ensemble agreement as a deferral condition, which can be applied directly to these models without any need for fine-tuning or specialized training.

\section{Agreement-Based Cascading (\coe)}
\label{sec:abchighlevel}

In this section, we present a high-level overview of the Agreement-Based Cascading
(\coe) approach, providing the essential concepts and insights needed to
understand \coe’s effectiveness in the experiments (\S\ref{sec:exps}).
Readers interested in the formalizations and technical details can refer to
\S\ref{sec:four} for a more comprehensive treatment.

\subsection{Overview of \coe Approach }

As outlined in Section~1, the goal of adaptive inference is
to identify data samples that can be evaluated accurately by a relatively
inexpensive model. This way, we can reduce the inference cost based on
whether a sample is `easy' or `hard'. A common approach to
this problem is deferral-based cascading, where we cascade over a sequence of models,
starting from the least expensive and using a `deferral rule' to determine if a
higher tier model must be used. These deferral rules are typically significantly
cheaper to evaluate (e.g. a small neural network), and add very little
additional cost on top of model execution. This 
approach makes it so that `simpler' cases are managed by smaller, faster models,
while only the complex cases cascade up multiple tiers and to more
resource-intensive models.

\textit{Agreement-Based Cascading (\coe)}, described in Algorithm~\ref{alg:cascade}, is one 
such deferral based cascading approach.
\coe maintains a set of
ensembles $\set{H_1, \dots, H_{n_E}}$ that starts from an ensemble of
inexpensive models in $H_1$ to expensive, state-of-the-art models in
$H_{n_E}$. Similar to other cascading techniques, given a sample $x$, we
start inference from the lowest (cheapest) tier in the cascade and use a
deferral rule, $r_i(x)$, to determine if a higher tier model is needed.
A core distinction between \coe and existing cascading approaches is its
agreement-based deferral rule. Instead of training a small additional deferral
network or post-hoc deferral rules, we use a notion
of \emph{agreement} between models within each  ensemble as a confidence
measure. When a (configurable) fraction of models in a tier agree, it signals that
they likely have the right answer. Conversely, if they disagree, the uncertainty
triggers a deferral to the next, more powerful ensemble. Additionally, we focus on cases where we wish to 
maintain accuracy of the overall prediction  and only improve inference cost
when accuracy does not suffer. This is different from existing approaches
that prioritize a fixed inference budget, potentially at the cost of accuracy.

{
\let\AND\undefined  
\setlength{\textfloatsep}{0pt}
\begin{algorithm}[t]
\caption{Agreement-Based Cascading (\coe)}\label{alg:cascade}
\begin{algorithmic}[1]
\REQUIRE  \updated{ Set of ensembles $\{H_1, H_2, \dots, H_{n_E}\}$, deferral
rule $r_i$ for each ensemble $i \in[n_E]$ as in Equation \ref{eq:voterules1} or \ref{eq:voterules2}}
\REQUIRE A new inference data point $x$.
\STATE Current cascade level, $i \gets 1$
\STATE Cascaded prediction, $y \gets \emptyset$
\FOR{$i \in \set{1, \dots, n_E}$}
   \STATE $y \gets H_i(x)$ 
\IF{$r_i(x)=0$} 
    \STATE \textbf{break} \hfill \COMMENT{Models in ensemble `agree'}
\ENDIF
\ENDFOR
\STATE \textbf{return} $y$
\end{algorithmic}
\end{algorithm}
}

%

\subsection{\coe in Modern ML Environments}

In recent years, obtaining trained models has become much easier, thanks to the
rise of public and private model repositories~\citep{wolf2019huggingface}, where a wide variety of
pre-trained models are readily available. This abundance of accessible models of
various sizes and performance levels makes the practical application of \coe
especially appealing.
Unlike
methods that require additional task-specific training or fine-tuning, \coe
 can leverage these existing models, making it suitable for direct deployment as a
``drop-in'' replacement for high-cost models. This adaptability, combined with
minimal setup and training requirements, can allow \coe~to fit seamlessly into many
existing ML workflows.

Strictly speaking, evaluating agreement between multiple models in an ensemble
is expensive when compared to the small router models that are used in many
existing approaches. However, two key aspects of modern ML workloads can help to mitigate
this additional cost in practice. First, in many cases some degree of
\emph{parallelization} is available that can reduce the impact on
inference cost metrics such as latency. We use $\rho$ to smoothly interpolate
between the fully sequential case $(\rho=0)$ to the fully parallel case,
$(\rho=1)$ as detailed in \S\ref{sec:safeagreement}. Second, in many
use-cases  the difference in cost between models in successive tiers of cascades
is so large that the impact of lower tier models on the overall cost of
inference is negligible, even with the added cost of constructing ensembles (see Figure~\ref{fig:syntheticparallel}).
We use $\gamma$ to denote the \emph{relative cost} of the models --- the ratio
of the cost of the smaller model to the larger model.

In \S\ref{sec:exps}, we evaluate \coe’s performance across several real-world scenarios (resulting in varying settings of $\rho$ and $\gamma$),
demonstrating its competitiveness and efficiency. We first show that \coe's
ensemble-based deferral mechanism preserves and often
improves on the accuracy of models it aims to replace. We also show that when the
lower tier models are small enough compared to the higher tier models, 
{\coe’s} ensemble-based
deferral rule operates with negligible impact on the overall cost.

We then evaluate cases in real-world workloads where such large relative costs
between various tiers of \coe is natural. In edge-to-cloud setups, \coe enables
substantial reductions in communication costs by processing simple tasks locally
on the edge. This minimizes the need to transfer data to the cloud, reducing
both latency and data transfer expenses. When serving models on heterogeneous
GPU resources in the cloud, \coe significantly reduces inference costs. By
selecting models based on task complexity, \coe makes efficient use of available
GPU resources, optimizing for both accuracy and rental cost. In black-box
API-based model deployments, where the user is billed per request or token, \coe  
offers substantial savings by reducing the average cost per request. By
deferring to high-cost models only when necessary, \coe achieves notable economic
savings compared to standard inference approaches.

{
\section{Formalization and Theoretical Details}
\label{sec:formal}
\label{sec:four}

\updated{In this section, we formalize our problem setup by first
outlining the standard statistical learning framework, which serves to define
the concepts of models, ensembles, the learning performance (e.g., accuracy) and
the inference cost associated with ensembles (\S\ref{sec:probsetup}).
Next, we describe deferral rule-based cascades and formulate the specific
case of \emph{drop-in cascades}, cascades which prioritise accuracy over
inference cost savings (\S\ref{sec:defcascade}). 
We introduce \emph{safe deferral rules} as a sufficiency condition for
constructing drop-in cascades and define our deferral rule based on 
agreement between models in an ensemble
(\S\ref{sec:safeagreement}). We conclude this section by formalizing the
test-time accuracy-inference cost behaviour of drop-in cascades (\S\ref{sec:coecost}). }




\subsection{Problem Setup and Notation}
\label{sec:probsetup}

\updated{Consider the standard statistical learning settting where 
$\mcX$ denotes an instance space and $\mcY$ denotes the label or response space.
Let $h(x)$ be a model taking inputs from $\mcX$ and producing outputs in
$\mcY$.  We assume that the models are from some hypothesis class $\mcH$. In
this setup, we typically characterize the learning performance of various models using
its \emph{risk} with respect to some data distribution $\mcD$ over $\mcX\times
\mcY$ and a loss function $l$, given by,
\begin{align*}
    \risk_{l}(h) = \E_{(x, y) \sim \mcD}[l(h(x), y)].
\end{align*}
As a concrete example, in the classification setup, where we use the
mis-classification error as a loss function, the risk is given by $ \risk(h) =
\E(y \ne h(x))$.}

Assume that each model $h \in \mcH$ has a cost of inference, denoted by
$C: \mcH \rightarrow \R_{+}$; for instance, for cases where we are concerned
about inference latency, the cost can be the latency of the model on the target
hardware.  Let $h_1$ and $h_2$ be two models with $h_2$ being the more expensive
of the two. We denote the \emph{relative cost} of the models by $\gamma :=
\frac{C(h_1)}{C(h_2)}$, satisfying $0 < \gamma \le 1$. 

Let $H^k: \mcX\rightarrow \mcY$ denote an ensemble of $k$ models from the
same hypothesis class $\mcH$. Similarly to before, the learning performance of
various ensembles can be characterized by their risk; let $\risk(H^k)$ denote
the risk of an ensemble $H^k$. Compared to a single member model, the cost of
evaluating the entire ensemble of models depends on
the various factors, including the degree of parallelization. Assuming that the
models in an ensemble are of similar cost, say $c_0$, we model the cost of the
ensemble using a \emph{parallelism coefficient} $0 \le \rho \le 1$ as,
\begin{align}\label{eq:costmodel}
    C(H^k) = c_0 k^{1-\rho} .
\end{align}
Here, when $\rho=1$, the ensemble suffers the same cost as a single model and
corresponds to the case where models can be fully parallelized. On the other
extreme, at $\rho=0$, the cost of the ensemble of $k$ models is $kc_0$,
corresponding to no parallelization (sequential evaluation).

\subsection{Deferral Rule-Based Drop-in Cascade}
\label{sec:defcascade}
 
A deferral-based cascade consists of a finite set of models  and a deferral
rule. Here, the idea is to start with the most resource-efficient model and use
the {deferral rule} to determine if a better-performing model with a higher
resource cost should be used for the current sample. The deferral rules
themselves are designed to have negligible inference cost given the models
inference outputs.  For this exposition, we will restrict ourselves to cascades
with only two levels, though the discussion in this section can be readily
generalized to larger cascades. Let the cascade consist of an ensemble $H^k_1$
at the lower level, and the larger model $h_2$ at the higher level;
$\mcM=\set{H^k_1, h_2}$. Let $r(x)$ denote a deferral rule;
$$
    r(x) = \begin{cases}
        1 & \text{defer to }h_2\\
        0 & \text{use }H^k_1
    .\end{cases}
$$
%
In this work, we focus on deployment scenarios where a decrease in model performance is a critical concern.
This is particularly relevant when cascades are intended to function as a
{drop-in replacement} for an existing deployment of an expensive model.
This notion can be formulated as maximizing the number of
calls to the smaller model while retaining the accuracy of the larger model.
With a slight abuse of notation, let $\mcM_r(x)$ denote the prediction of the
cascade with $\mcM = \set{H^k_1, h_2}$ and deferral rule $r$. Then we desire that for a choice of a small error budget $\xi > 0$,
\begin{align*}
    \max_{r}&\quad \prob(r(x) = 0)\\
    \text{s.t.}&\quad 
        \prob(y \ne \mcM_r(x)) \le \prob(y \ne h_2(x)) + \xi,
        \numberthis
        \label{eq:prog2}
\end{align*} 
The constant function $r(x) = 1$, which defers for every $x$, is feasible and
attains the objective value of $0$. Moreover, \emph{every} feasible deferral
rule leads to a cascade with competitive accuracy as $h_2$.  Instead of picking
an error budget $\xi$, a complimentary (dual) approach is to consider a fixed
inference budget and aim to attain the best accuracy within the budget. In this
view,  drops in accuracy (compared to the existing large model) are permissible,
provided the inference budget is met. 
 ~\citet{jitkrittum_when_2023} examine this perspective, which we recommend to
 interested readers.




\subsection{Deferral using Ensemble Agreement}
\label{sec:safeagreement}

Ensemble agreement is an instantiation of score based deferral
rules~\citep{jitkrittum_when_2023}. In its simplest form, we associate a score to
the prediction output of an ensemble model and interpret this score as a
measure of confidence in that particular prediction. We can then defer to a
larger model when the confidence is below a predetermined threshold. For a
model $H_1^k(x)$ and a scoring function $s(x)$, the deferral rule is defined as
\[
    r(x) = \eye[s(x) \le \theta],
\] where $\eye$ is the indicator function. 
In general, such score based deferral rules can sometimes be misleading.  For
instance, in multi-class classification, rules that map the outputs of a
classifier to a probability distribution over labels can produce confidently
incorrect predictions when an unperceivable amount of perturbation is added to
the data sample~\citep{szegedy2013intriguing}. However, such rules have been
shown to be effective in practice~\citep{wang_idk_2018, gangrade_selective_2021,
mamou_tangobert_2022, gupta2024language} implying that such adversarial data is
rare in many cases. Motivated by this observation, we propose the following
property of scoring functions (and by extension the corresponding deferral
rule).

\begin{definition}[Safe deferral rule] Let $H^k: \mcX\rightarrow \mcY$, $k\ge1$ be a
classifier and let $s: \mcX \rightarrow [0, 1]$ be a scoring function. The
scoring function $s$ is referred to as a safe scoring function for $H^k$ if there exists $\theta \in [0, 1]$ such that, for some small $\eps > 0$,
\[
    \prob(s(x) \ge \theta, H^k(x) \ne y) \le \eps.
\] The corresponding deferral rule $r(x) = \eye[s(x) \le \theta]$, is referred to as a safe deferral rule, and satisfies $ \prob(r(x)=0, H^k(x) \ne y) \le \eps. $
\label{def:safe}
\end{definition} 
We define safe deferral with respect to classification tasks for simplicity. The
definition can be extended to other cases with appropriate choice of loss function.
Intuitively, this formalizes the notion that the deferral rule can  
probabilistically identify a \emph{subset} of $\mcX \times \mcY$ for which
$H^k_1(x)$ is correct. Alternatively, we can think of the deferral rule as a one
sided classifier, similar to the problem studied in \cite{goyal2020drocc}, where
the class to be identified is the \emph{subset} of $\mcX \times \mcY$ where
$H^k_1(x)$ is correct. In general, such a safe deferral rule need not exist for
a given model $H_1^k(x)$. However, in \S\ref{sec:exps} we demonstrate
empirically that such rules can in-fact be constructed without additional
training for many real world tasks with appropriate choice of $\theta$
(Figure~\ref{fig:selection_rate}). These rules lead to selection rates as high
as 90\% of the data in cases such as ImageNet-1K.

Assuming access to a safe deferral rule, observe that for $\xi \ge \eps > 0$,
such rules are feasible for the program in Equation~\ref{eq:prog2} (see
Proposition~\ref{prop:costacc}). This implies that \emph{every} cascade, $\mcM =
\set{H_1^k, h_2}$ using a safe deferral rule $r$, is competitive with the larger
model $h_2$ in terms of accuracy. We empirically evaluate these drop-in cascades
in \S\ref{sec:exps}. Note that the optimal safe deferral rule can lead to
an improvement in the accuracy of \coe in theory (see,
Appendix~\ref{app:admissible}), and we see accuracy improvements in our
experiments (\S\ref{sec:exps}).


\vspace{-2mm}
\paragraph{Agreement-based deferral rule.} We evaluate two
flavors of deferral rules that capture the notion of agreement between models.
In cases where we have direct access to the models, we directly use the outputs
produced by models in the ensemble. However, in certain cases, for instance when
interfacing through a third-party provider's inference API, we only have
black box access to models. For such cases, we use a voting scheme between the
models in an ensemble to construct our scoring function. Concretely,  for an
ensemble $H^k$ consisting of $k \ge 1$ models, let $s(x; H^k)$ denote
the average score of the majority prediction on $x$ and let $\vote(x; H^k) =
\frac{1}{k}\sum_{h \in H^k} \one [H^k(x) = h(x)]$ denote the fraction of votes
received by the prediction, where,
\begin{align}
r_{\text{v}}(x; \theta_{v}) &= \begin{cases}
    1  & \vote(x; H^k_1) \le \theta_v \\
    0 & \text{otherwise}.
    \label{eq:voterules1}
\end{cases}
\\
r_{\text{s}}(x; \theta_{s}) &= \begin{cases}
    1  & s(x; H^k_1) \le \theta_s \\
    0 & \text{otherwise}.
    \label{eq:voterules2}
\end{cases}
\end{align}




\subsection{Inference Cost Savings and Competitiveness}
\label{sec:coecost}

Since we do not impose an inference cost budget on the cascade, it is possible
for the cascade to incur a higher cost than simply using the larger model $h_2$.
Intuitively, if majority of the samples seen during test time are `hard', the
cascade pays the cost of evaluating both $H^k_1(x)$ and $h_2(x)$.

\begin{proposition}\label{prop:costacc} Let $\mcM = \set{H_1^k, h_2}$ be two classifiers and $r$ a deferral rule such that $r$ is a safe deferral rule for $H_1^k$ according to Definition~\ref{def:safe}, for a distribution $\prob$ over $\mcX \times \mcY$. Then for every $\xi \ge \eps > 0$, the agreement based cascading (\coe) classifier $\mcM_r(x)$ is such that,
\begin{enumerate}
\item The \coe classifier is competitive with the large classifier $h_2$ in terms of accuracy (zero-one loss),
    \[ \risk(\mcM_r)  
    \le 
    \risk(h_2) + \eps, \]
\item The \coe classifier enjoys an average inference cost of,
\[ \E[C(\mcM_r)] = (k^\rho\gamma + \prob(r(x) = 1))C(h_2). \]
\end{enumerate}
\end{proposition}
The proof follows directly from Definition~\ref{def:safe} and basic probability (see Appendix~\ref{app:admissible}). Safe deferral rules can be constructed with minimal cost: our threshold estimation requires only $\sim$100 validation samples and simple voting computations, avoiding expensive router training.
Thus, the cost savings we can expect with a drop-in cascade depend on three key
factors: the relative cost $\gamma$, the degree of parallelization $\rho$ and the
deferral rate  $P(r(x) = 1)$, or equivalently, the \emph{selection rate}
$P(r(x) = 0)$. In particular, for \emph{every} feasible deferral rule $r$,
in the best case scenario where the cost of the smaller model is negligible
(i.e., $\gamma = 0$), the cost of inference reduces by the selection rate, $P(r(x) =
0)$. Conversely, in the worst-case scenarios, the cost can be $(k+1)$ times the
cost of the larger model. In the next section, we demonstrate real world
scenarios where the favorable interplay of these three quantities lead to
significant improvements in inference cost. \updated{See
Appendix~\ref{app:selection} and Figure~\ref{fig:selection_rate} for information
on selection rates for various models and datasets considered here.}

}

\begin{figure*}[t!]
  \begin{subfigure}[t]{0.24\textwidth}
  \centering
    \includegraphics[width=\textwidth]{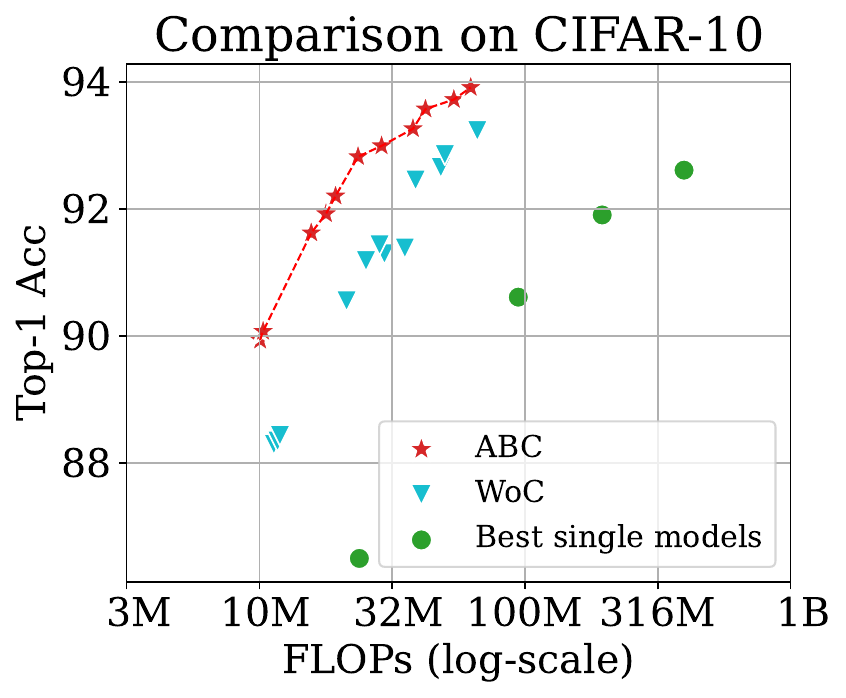}
  \end{subfigure}
  \begin{subfigure}[t]{0.24\textwidth}
  \centering
    \includegraphics[width=\textwidth]{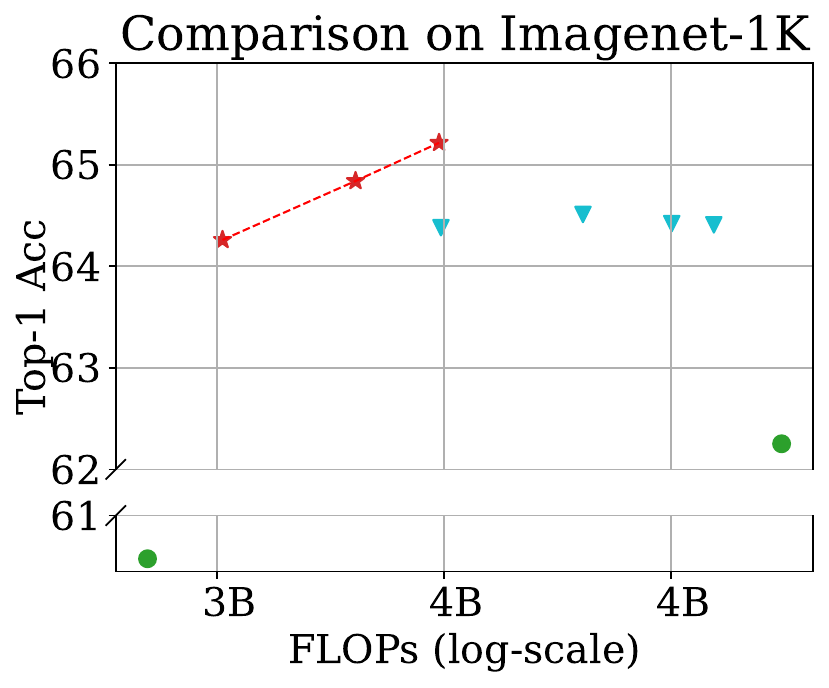}
  \end{subfigure}
  \begin{subfigure}[t]{0.24\textwidth}
  \centering
    \includegraphics[width=\textwidth]{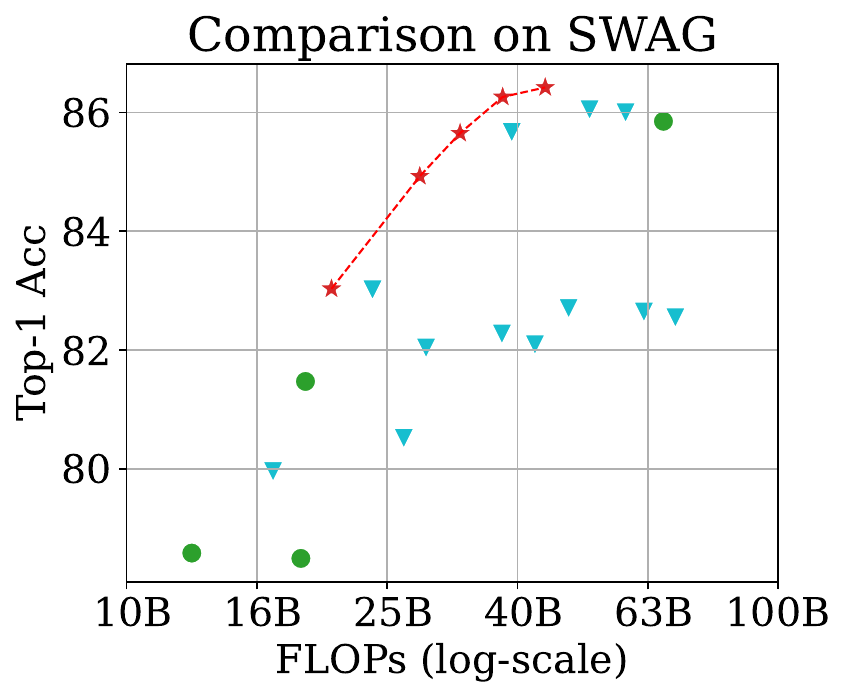}
  \end{subfigure}
  \begin{subfigure}[t]{0.24\textwidth}
  \centering
    \includegraphics[width=\textwidth]{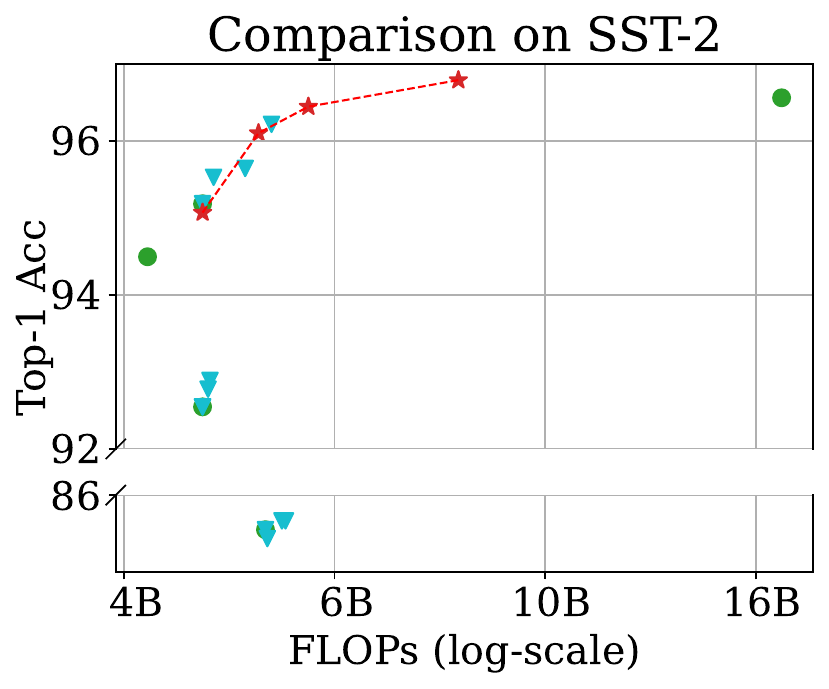}
  \end{subfigure}
  \caption{Pareto curves of \coe
  vs. confidence-based cascades (WoC) \citep{wang2021wisdom} and best single models on diverse tasks. 
  For WoC, we tune its cascade configurations across the best four of its
  confidence thresholds and generate results from their most performant
  cascades. \coe maintains a Pareto-optimal curve, which consistently
  outperforms both methods in accuracy with lower FLOPs costs.
  }
  \label{fig:approach_comparison}
  \vspace{-3mm}
\end{figure*}
\section{Experiments}
\label{sec:exps}

This section evaluates our training-free Agreement-Based Cascading (\coe)
approach across a variety of language and vision tasks, focusing on accuracy and
inference cost. First, in \S\ref{sec:freeparallel}, we examine \coe's accuracy-cost tradeoff against state-of-the-art models under full parallelization, setting the \emph{parallelization factor} $(\rho)$ to $\rho=1.0$  
for optimal tradeoff. Next,
in \S\ref{sec:synthetic}, 
we show that using ensembles at lower tiers has minimal impact on inference cost when lower-tier models are at least $50\times$ cheaper than the largest model. In such cases, the \emph{relative cost} $(\gamma)$ satisfies $\gamma \le \frac{1}{50}$, making \coe effective without parallelization.
Finally, \S\ref{sec:real_world} explores practical scenarios where low
relative costs 
make
\coe is naturally suited for real-world
deployments.


\paragraph{Estimating Voting Threshold:}
\coe's deferral rule uses a configurable voting threshold, $\theta$ (see
Equations~\ref{eq:voterules1} and~\ref{eq:voterules2}) at each cascading tier.
We estimate $\theta$ empirically on a small set of unseen data; see App.~\ref{app:samplecomplexity} for details.


\paragraph{Datasets:} %
To evaluate \coe, we use a range of benchmark datasets for image and language tasks, as shown in Table \ref{tab:datasets_used} in the Appendix. Additional datasets are used in \S\ref{sec:api_settings} to align with those explored by state-of-the-art baselines. 

\vspace{-2mm}
\paragraph{Models:} We select diverse models for both image and language tasks, summarized in Appendix's Table \ref{tab:models_used}. For BERT-based models, we use the \texttt{BASE} and \texttt{LARGE}, and for image models, we tier by FLOPs count. All models are sourced from HuggingFace Zoo for inference without any additional training effort on our end. \S\ref{sec:api_settings} uses models from LLaMA~3, Gemma~2, and Qwen~2 families via the Together API. We detail this section's experimental setup in App.~\ref{app:methodspecific}. 

\paragraph{Evaluation:} For the generation tasks we consider in \S\ref{sec:api_settings}, the datasets each consist of a fixed set of possible outputs
akin to the classification tasks, and we apply our deferral rule on the final output at each tier. Certain cases don't have a fixed set of output labels (while still not being open-ended generation), like (1) GSM8K, where there is a final numeric answer at the end of the generated output to the math word problems, and (2) CoQA, where we used F1-score to capture overlaps between predictions and ground-truth answers. We do not discuss open-ended generation tasks in this work as the deferral rule we consider—and the baseline methods that we compare to---is not directly applicable to such cases.

\subsection{When is~\coe Practical?}

\subsubsection{When Parallelization is Cheap}
\label{sec:freeparallel}
We first consider the case  where the inference cost of an ensemble of models is
the same as the cost of a single model. This idealistic scenario could happen, for instance, in offline batch inference~\citep{aws_batch_2021}, when GPUs are available for parallelization, or the models are small enough that existing resources are under-utilized and an ensemble adds no additional cost.
More importantly, this setting establishes best-case accuracy and inference cost
values for \coe. We consider total floating-point operations (FLOPs) as a
representative inference cost metric here, and discuss other metrics such as
communication cost, latency, or cloud rental costs in subsequent subsections.

\begin{wrapfigure}{r}{0.53\textwidth}
\vspace{-7mm}
\centering
  \centering
    \includegraphics[width=0.53\textwidth]{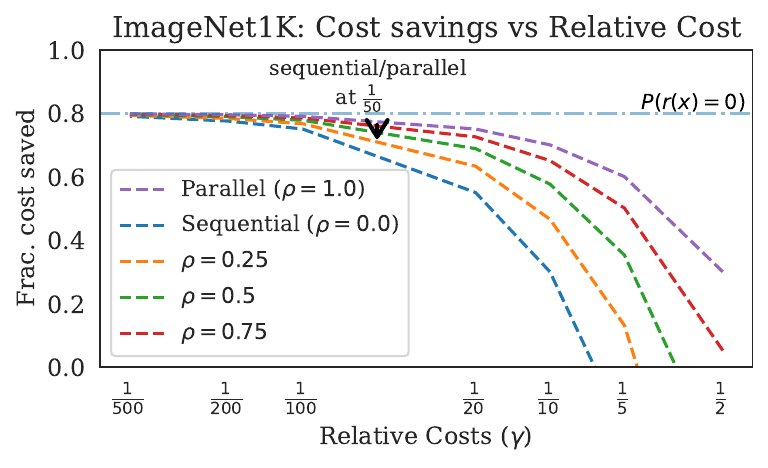}
  \vspace{-7mm}
  \caption{
      Fraction of inference cost saved as a function of relative 
      cost of models $(\gamma)$, \emph{assuming} a fixed selection rate
      $\prob(r(x) = 0)$. As parallelization decreases from fully parallel
      $(\rho=1)$ to sequential $(\rho=0)$, cost of evaluating ensembles increase
      and the cost savings decrease. When models across tiers are of similar
      size (e.g., smaller model is at most $5\times$ smaller, $\gamma \ge
      \frac{1}{5}$), some parallelization is needed for \coe to reduce costs
      effectively. However, for lower relative costs (e.g., smaller model is at
      least $50\times$ smaller, $\gamma \le \frac{1}{50}$), sequential and
      parallel settings achieve meaningful savings, showing \coe’s efficiency
      even with the added cost of using ensembles.
  }
  \label{fig:syntheticparallel}
  \vspace{-8mm}
\end{wrapfigure}
The accuracy vs. FLOPs for \coe is shown in Figure~\ref{fig:approach_comparison}. As a point of comparison, we also include Wisdom-of-Committees (WoC) \citep{wang2021wisdom}, a popular and representative confidence-based model cascading method.  In most cases, we observe that \coe is able to improve the Pareto frontiers, as it usually sees a $1$--$2$ point increase in accuracy. We attribute this to a combination of (a) the improvement ensembles can have on accuracy that is widely noted in literature~\citep{gontijo-lopes_no_2022, jiang_llm-blender_2023} and (2) the improvement agreement-based rules can cause in
cascading (Appendix~\ref{app:admissible}). Overall, in terms of accuracy, \coe
either exceeds---or at least matches---the accuracy of the best models
at a fixed FLOPs budget, and is a practical {drop-in} replacement for these
models.

\subsubsection{Disparity in Relative Cost is High}
\label{sec:synthetic}

Of course, in many real-world scenarios, the cost of evaluating ensembles
is not negligible. For instance, in the edge-to-cloud inference scenario discussed in \S\ref{sec:edgetocloud}, models in an ensemble often are evaluated sequentially. However, even in such scenarios, if the relative cost of models across each level of the cascade, $\gamma$, is small enough, this additional cost becomes negligible compared to the overall cost.



In Figure~\ref{fig:syntheticparallel}, we demonstrate the impact 
of using ensembles at various relative costs on the overall inference cost. As
shown in the first plot, as we move away from the ideal, fully parallel setting
with $\rho=1$ towards the sequential setting with $\rho=0$, the fraction of
inference cost saved decreases. In fact, when the models in the cascade are
of similar size---for example, when $\gamma \ge \frac{1}{5}$---a certain
degree of parallelization is required for \coe~ to reduce inference
cost. However, observe that when the relative costs are small enough (e.g., $\gamma \le \frac{1}{10}$), the need for parallelization diminishes. This is
evident in Figure~\ref{fig:syntheticparallel} (right); for $\gamma \le
\frac{1}{50}$, the sequential execution curve $(\rho=0)$ approaches the curve with
full parallelization $(\rho=1)$.

\begin{AIbox}{Takeaway \#1: }
\begin{itemize}[leftmargin=0.7em]
    \setlength\itemsep{0em}
    \item  Although ensembling requires additional inference costs, these costs can be mitigated when (1) models can be parallelized or (2) smaller models are several magnitudes cheaper than larger ones. In such scenarios, \coe can achieve substantial accuracy improvements while reducing inference costs.
\end{itemize}
\end{AIbox}

\subsection{Real-world Use-cases}
\label{sec:real_world}
As noted in the previous two subsections, \coe~ either improves on, or is at
least competitive with, the single best model in terms of accuracy. Moreover,
whenever the relative cost is small enough, \coe~ suffers negligible additional penalty
for using an ensemble of models. As noted in Figure~\ref{fig:pareto}, accuracy
vs inference cost scaling for ML models already imply that the relative cost of
models that only differ a few points in accuracy is small. For instance,
the state-of-the-art model performance on the ImageNet-1K dataset attains about
$83\%$ top-1 accuracy with $70$B FLOPs. A Pareto-optimal model that
achieves $63\%$ accuracy requires only about $1$B flops, and thus a two-level \coe with ensembles of these two models has a relative cost of $\gamma =
\frac{1}{70}$.

In many real world scenarios, this disparity in relative cost is further amplified due
to deployment considerations --- for instance, in an edge-to-cloud
inference setting, local inter-process communication (IPC) latency is typically
two orders smaller than remote (cloud) IPC latency $(\gamma \approx 10^{-2})$. Moreover,  the ability of
models in~\coe to be placed on multiple, distributed devices with negligible
synchronization overhead adds more opportunities for inference cost reduction.
We consider three such scenarios, their respective cost models, and the benefits of
\coe in these cases here.
These include edge-to-cloud inference (\S\ref{sec:edgetocloud}), cloud-based model serving on heterogeneous GPUs (\S\ref{sec:gpu_costs}), and black-box inference access to model API services (\S\ref{sec:api_settings}).


\begin{figure*}[t!]
    \begin{subfigure}{0.5\textwidth}
    \centering
      \includegraphics[width=0.96\textwidth]{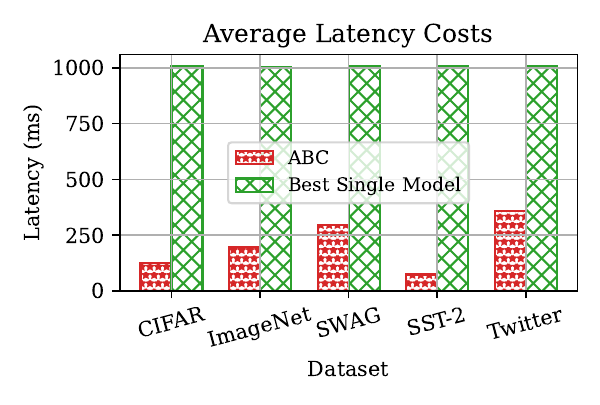}
    \end{subfigure}
    \begin{subfigure}{0.5\textwidth}
    \centering
      \includegraphics[width=0.95\textwidth]{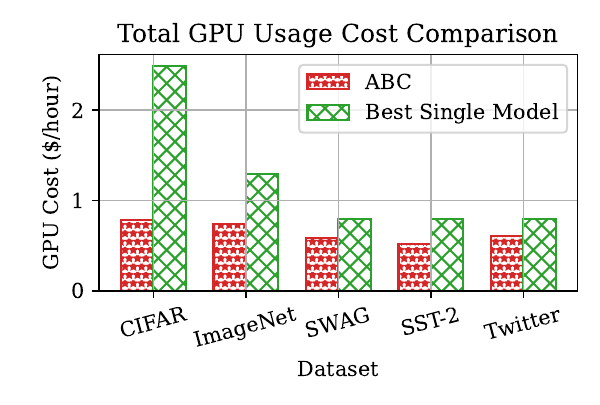}
    \end{subfigure}
    \vspace{-5mm}
\caption{(a) \coe for edge-to-cloud inference: We simulate a single-instance
inference setup, as seen in real-time applications where predictions may need to
be made as new data becomes available. \coe can enable small models to be served
at the edge without sacrificing accuracy---leading to large savings in
communication costs over the alternative of using only the highest
accuracy/largest model residing in the cloud, or a single small and
low-performing model on the edge. (b) Total GPU usage costs of \coe vs. using
the best model. Agreement-Based Cascading, at reduced costs of GPU usage, exceeds the
accuracy of the single best models in all task categories.
}
    \label{fig:gpu_costs}
\end{figure*}

\subsubsection{Communication Cost in Edge-to-Cloud Inference}
\label{sec:edgetocloud}

An advantage of \coe is that it allows a single large model to be split
into multiple, potentially much smaller models, with only a simple reduce
operation required to compute agreement.  This allows us to tune device
placement at various levels of \coe~to improve inference costs. One use-case
where this is beneficial is edge-to-cloud inference~\citep{forooghifar2019resource}; here,
inference requests are generated on user-facing edge devices like mobile phones
or smart devices, which are sent over the network to a cloud service for
evaluation. Given an inference request generated by a user interaction, the
time-to-response or response latency in such case is dominated by communication
overheads (network speed, serialization overhead, network congestion, etc.)
beyond our control.  By using \coe for such applications, we are able to
distribute the inference load between tiny, on-device models and the cloud
models, allowing us to avoid communication costs for a significant portion of requests.

To understand the effectiveness of \coe in such a scenario, we consider a
communication cost model previously studied in \citet{zhu2021delayed,
lai2022fedscale} in a setup of edge devices (i.e., Raspberry Pis and
smartphones) and cloud servers. The delay parameters adopted range from small,
medium, to large [1~us, 10~ms, 100~ms, 1000~ms], where near-instantaneous local
communication (\textless~1 microsecond) can be expected to occur with base
cascade tiers performing inference on-device, and substantial network delays
might occur (\textgreater~1 second) in a worst-case edge-to-cloud transmission
($\gamma = 10^{-6}$).We simulate this by considering a two-level
cascade, with the smaller level placed on the edge-device. We apply the delay to
the cascade exit points on the edge device to capturing the time cost of
transitioning between edge-to-cloud.

Our results, as shown in Figure \ref{fig:gpu_costs}, show that the
flexibility that \coe affords in terms of model placement allows significant latency
reductions, while providing superior accuracy compared to a single cloud model.
In particular, we see that cascading in these scenarios provides an ~14$\times$ reduction in communication cost for language tasks like SST-2; and for image datasets, we see a $5\times$ reduction for ImageNet-1K and a $8\times$ reduction in CIFAR10.

\begin{AIbox}{Takeaway \#2: }
\begin{itemize}[leftmargin=0.7em]
    \setlength\itemsep{0em}
    \item  \coe enables model placement flexibility where small models run locally and large models in the cloud. Communication delays can make relative costs significant, resulting in \coe achieving 5-14$\times$ reductions in communication costs while maintaining superior accuracy.

\end{itemize}
\end{AIbox}




\begin{figure*}[t]
    \vspace{-10mm}
    \centering
    \begin{subfigure}{0.45\textwidth}
    \centering
      \includegraphics[width=\textwidth]{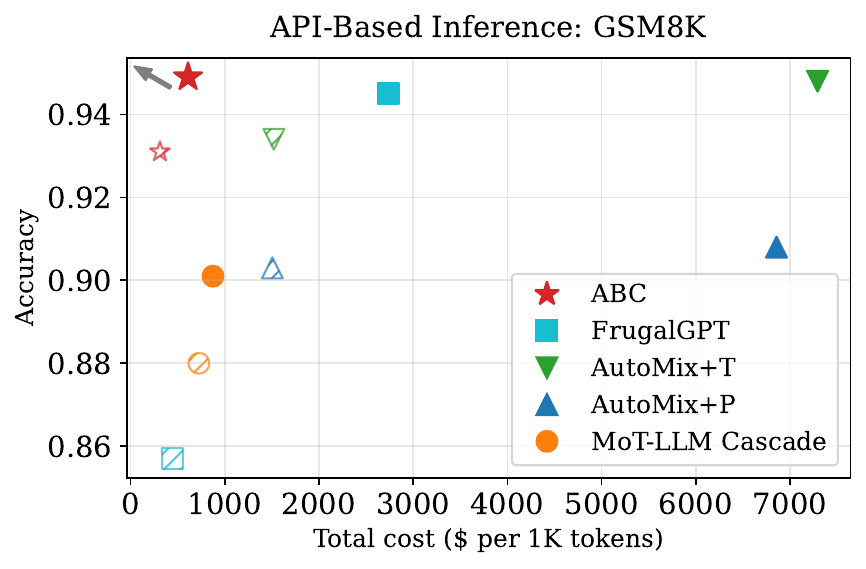}
    \end{subfigure}
    \begin{subfigure}{0.45\textwidth}
    \centering
      \includegraphics[width=\textwidth]{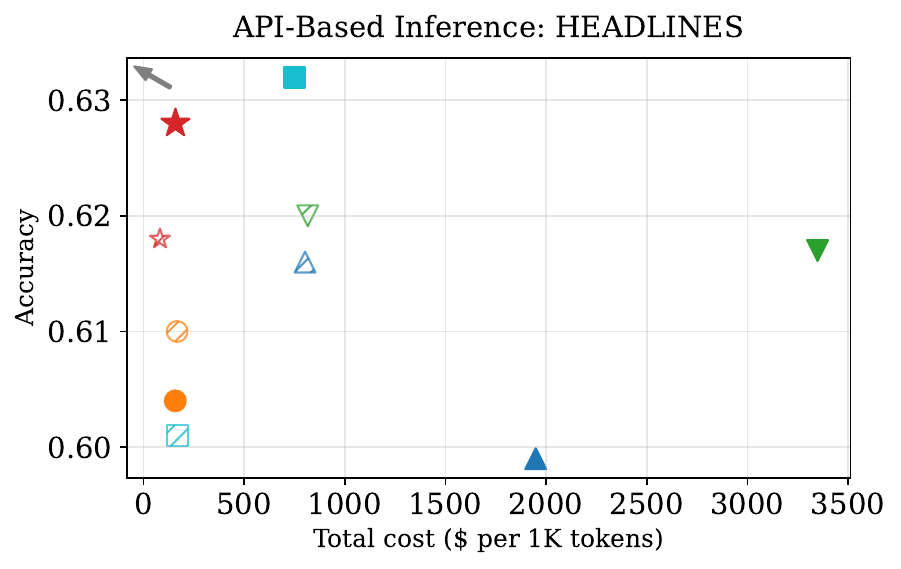}
    \end{subfigure}\hfill
    \begin{subfigure}{0.45\textwidth}
    \centering
      \includegraphics[width=\textwidth]{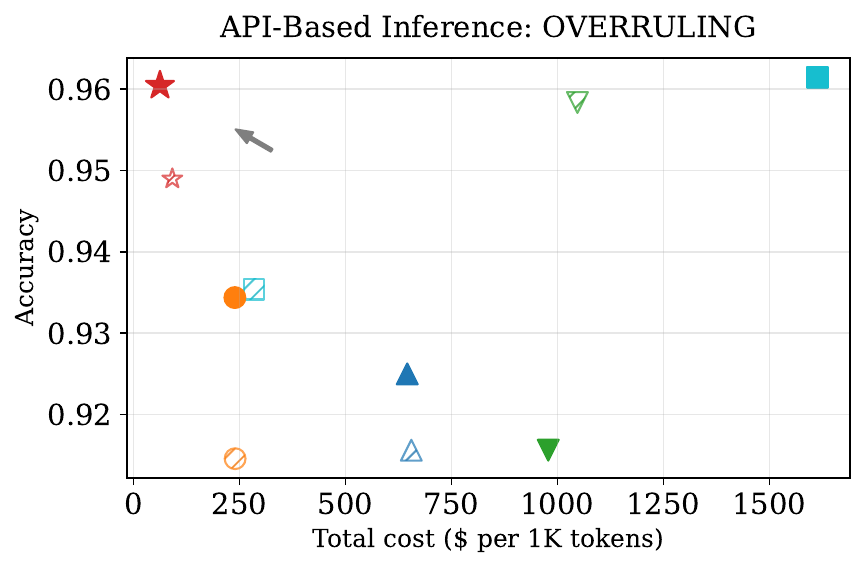}
    \end{subfigure}
    \begin{subfigure}{0.45\textwidth}
    \centering
      \includegraphics[width=\textwidth]{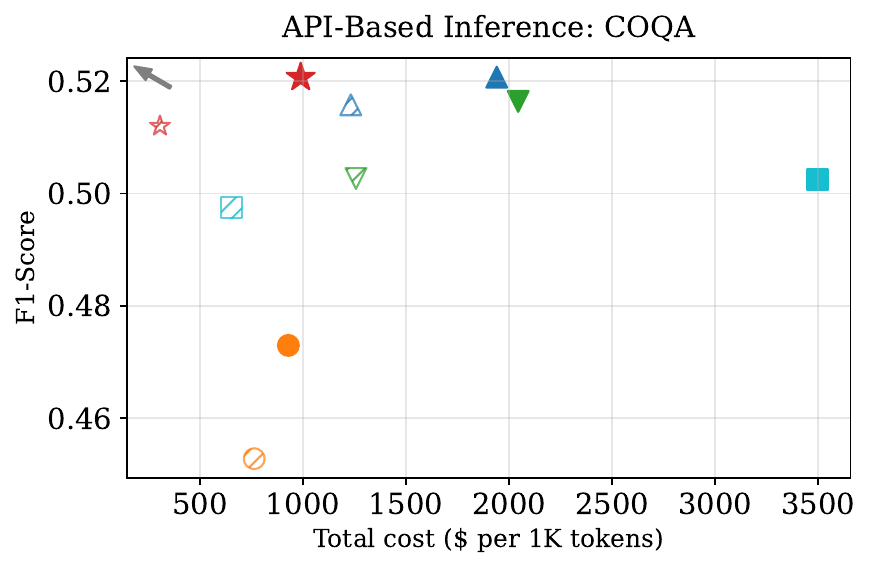}
    \end{subfigure}
    \vspace{-.1in}
\caption{Comparison of \coe against state-of-the-art cascade baselines for
black-box API-based inference. The faded, hatched-patterned variants represent
budget-friendly, 2-level cascade instances where we do not include the costly
Tier 3. Most of these methods show competitive performance, but \coe matches
their accuracy at significantly lower costs in all tasks. Note that all these
methods (aside from the MoT-LLM cascade) incur additional setup costs not
reflected in our plots.
}
    \label{fig:api_plots}
    \vspace{-5mm}
\end{figure*}

\subsubsection{Monetary Cost for Model Serving on Heterogeneous Hardware}
\label{sec:gpu_costs}
Another use-case which can take advantage of the model placement flexibility of
\coe is using heterogeneous hardware for model serving on the
cloud~\citep{crago2015heterogeneous, li2021serving, li2023kairos, mo2023hetsev}.
GPU/Accelerator hardware typically has a disproportionately large difference in
hardware costs compared to their throughput difference. For instance, based on
the current pricing model offered by Lambda~\citep{lambdacloud}, a popular cloud
rental platform, the rental pricing of a single A100 is \$1.40/hour and a
V100 node is \$0.06/hour ($\gamma \approx 4\times10^{-2}$), while the rated
32-bit tensor core throughput is 312 TFLOPs for A100 and 125 TFLOPS for V100. In
this scenario, a simple placement strategy for a 2-level \coe~that reduces
inference cost may place the smaller model on V100 nodes and larger models
on A100 nodes. Since the lower level models are---at least---an order of magnitude
cheaper than the larger one in terms of FLOPs, this offsets the throughput loss
we incur when switching from A100 to V100; and as a result, the \coe~implementation
incurs a much lower average cost.




To concretely evaluate \coe under such a cost model, we retrieve the costs of GPU
usage by hour from Lambda Cloud's pricing (see, Table~\ref{tab:gpu_costs} in
the appendix for details). For simplicity, we assume that each ensemble tier is
set up on a distinct GPU in increasing order of GPU sophistication, and serves a
uniform inference request rate. We also assume that the nodes are co-located and
communication cost between them is negligible. We show a summary in
Figure~\ref{fig:gpu_costs} and present the detailed results in Table
\ref{tab:costs_table}, in the Appendix. We can see at least a $3\times$
reduction in inference cost in terms of $\$/$hour for image tasks, and a more
moderate $10$-$30\%$ reduction in language tasks. 

\begin{AIbox}{Takeaway \#3: }
\begin{itemize}[leftmargin=0.7em]
    \setlength\itemsep{0em}
    \item  GPU rental cost differences significantly exceed throughput differences. \coe optimally places models across hardware tiers, achieving 3$\times$ cost reductions for image tasks and 10-30\% savings for text tasks.
\end{itemize}
\end{AIbox}


\subsubsection{Monetary Cost for Black-box API-Based Inference}
\label{sec:api_settings}
\begin{wraptable}{r}{0.43\textwidth}
\small
\vspace{-1mm} 
  \centering
  \begin{tabular}{c c c}
    \toprule
    \textbf{Tier} & \textbf{Model} & Price \\
    \midrule
    \multirow{3}{*}{Tier 1} & LlaMA 3.1 8B-Instruct Turbo  & 0.18  \\
                            & Gemma 2 9B IT               & 0.30  \\
                            & LlaMA 3 8B Instruct Lite    & 0.10 \\
    \midrule
    \multirow{3}{*}{Tier 2} & LlaMA 3.1 70B Instruct Turbo & 0.88  \\
                            & Gemma 2 27B Instruct              & 0.8  \\
                            & Qwen 2 72B-Instruct          & 0.9 \\
    \midrule
    Tier 3                  & LlaMA 3.1 405B Instruct Turbo & 5.0 \\
    \bottomrule
  \end{tabular}
  \caption{The cascade tiers, their models, and associated costs, in
  dollar per million tokens, for the API-based experiments; all API services are
  provided by \href{https://api.together.ai/models}{together.ai}. We use the
  tiers and models as they are for \coe inference system. However, for the
  single-model cascade baselines, we use the tiers' best models
  for their systems.}
  \label{tab:model_costs}
\vspace{-5mm}
\end{wraptable}

Finally, in the current landscape dominated by LLMs, many providers offer black-box API
access to their proprietary LLMs~\citep{abdalla2023elephant, sun2022black, hadi2024large}. Similarly to the large cost difference
between GPU generations discussed in the previous subsection, we also observe a
large cost disparity between various generations/tiers of API calls.
For example, using Together.ai---one of the lowest-cost serverless endpoint providers for LLMs\footnote{\href{https://www.together.ai/pricing}{together.ai/pricing} as of September 2024.}---as a case study, we find that models within the 7B-8B range cost \$0.20 per million tokens, while LlaMA3.1-405B costs \$5.00. This translates to the larger model costing 25$\times$ more than the smaller model range $(\gamma = \frac{1}{25}, \rho=1)$. If we consider GPT-4 as the gold standard, the cost of usage quickly scales to 150$\times$ of our reference smaller model's costs.\footnote{\href{https://openai.com/api/pricing/}{GPT-4-1106-preview} costs \$30 as of September 2024.}.

Since we only have black-box access to these models, we cannot employ a score-based deferral rule. However, we demonstrate that using \coe's voting-based rule defined in \S\ref{sec:formal} is also effective in such scenarios. For baseline comparison, we use FrugalGPT, 2 variants of AutoMix, and MoT LLM Cascade. 
These state-of-the-art cascading methods are specifically designed for scenarios where we
make API calls to black-box model endpoints, in contrast to our more generally
applicable \coe method. 
We access the models from Together.ai---described in
Table~\ref{tab:model_costs}---for these experiments, and  consider a setting that
is advantageous to the baselines, by selecting the \textit{best singular model}
from each performance tier in their respective approaches. 

All the baseline methods considered here are significantly more complex and
involved to set up than \coe; both AutoMix and FrugalGPT involve training a
router or deferral rule at each cascade level which has to be repeated for every
new task or model change. Unlike AutoMix, FrugalGPT requires training a
DistilBERT-based scorer, which would require the user's possession of GPU
resources.
MoT LLM Cascade generates multiple results and varies the randomness in the LLM’s responses via sampling, while using in-context demonstrations and reasoning
techniques (e.g., Chain-of-Thought~\citep{wei2022chain}) to influence how the
model generates answers. 
In contrast, \coe~uses a much simpler voting-based safe deferral rule, without
involving additional training or complex routing strategies. Further method-specific details for these experiments as well as additional results can be
found in Appendix~\ref{app:methodspecific}.


Our results, as shown in Figure \ref{fig:api_plots}, demonstrate that \coe
is a more reliable deferral rule, and thus,  
offers a more favorable trade-off between accuracy and cost compared
to existing methods---despite
their more sophisticated routing mechanisms, and even without considering the 
additional setup costs incurred by some of the baseline methods. 
For instance, while FrugalGPT's scorer struggles on harder tasks and tends to take the safer route by deferring more frequently, \coe aggressively leverages cheaper models for a significant portion of inputs, reserving higher-cost models only when necessary. 
AutoMix, on the other hand,
uses a few-shot self-verification system that is sampled $k$ times where $k$ is \textgreater1
(in the authors' codebase and ours, $k=8$);
hence, the additional API calls add significantly to its cost of usage.
In contrast, \coe easily maintains and often improves on accuracy while being a
training-free, simple approach.

\begin{AIbox}{Takeaway \#4: }
\begin{itemize}[leftmargin=0.7em]
    \setlength\itemsep{0em}
    \item  LLM API pricing creates extreme cost ratios. \coe's simple voting mechanism achieves 2-25× cost reductions compared to SOTA cascading methods without requiring complex router training.
\end{itemize}
\end{AIbox}

\subsection{Ablation Studies and Sensitivity Analysis}
We conduct comprehensive ablation studies to validate \coe's design choices and analyze sensitivity to key deployment parameters:

\paragraph{Impact of Parallelization} \S\ref{sec:freeparallel} and Figure~\ref{fig:parallel} (in Appendix~\ref{app:sequential_parallel}) analyze \coe under different parallelization scenarios. With full parallelization ($\rho=1$), \coe achieves optimal accuracy-FLOPs trade-offs, consistently outperforming single models. Even under sequential execution ($\rho=0$), \coe maintains substantial advantages when relative costs are sufficiently disparate (e.g., $\gamma \leq \frac{1}{50}$), demonstrating robustness to deployment constraints.

\paragraph{Relative Cost Sensitivity}
\S\ref{sec:synthetic} and Figure~\ref{fig:syntheticparallel} examine how \coe depends on the relative cost $\gamma$ between cascade tiers. When models have similar costs ($\gamma \geq \frac{1}{5}$), some parallelization may be required for cost savings. However, when smaller models are at least 50$\times$ cheaper ($\gamma \leq \frac{1}{50}$), \coe provides meaningful savings even with sequential execution, validating our theoretical analysis.

\paragraph{Threshold Estimation Robustness}
Appendix~\ref{app:samplecomplexity} and Figure~\ref{fig:estimationrate} demonstrate that agreement threshold $\theta$ estimation is stable across different model accuracy levels (37.6\% to 86.0\%) and converges with minimal validation data. Using only 100 samples yields threshold estimates that remain stable when evaluated on 10$\times$ more data, confirming the effectiveness of the calibration procedure.

\paragraph{Safe Deferral Rule Existence}
Appendix~\ref{app:selection} and Figure~\ref{fig:selection_rate} characterize selection rates under different error tolerances (1\%, 3\%, 5\%) across model accuracies and computational budgets. Higher-accuracy models achieve selection rates up to 60\% even with strict 1\% error tolerance, empirically validating the practical existence of safe deferral rules predicted by our theory.

\paragraph{Cost Breakdown Analysis} Tables \ref{tab:gpu_costs} \& \ref{tab:costs_table} (Appendix~\ref{app:cost_analysis}) provide detailed analysis across cascade tiers, showing that most samples (52 to 93\%) are processed at cheaper early tiers. This validates \coe's ability to concentrate expensive computation on truly difficult samples while handling the majority of requests efficiently.

\paragraph{Cascade Configuration Effects} Figure~\ref{fig:parallel} explores different cascade lengths (2-4 levels) and ensemble sizes (2-5 models per tier). The results show diminishing returns for larger ensembles, with 2-3 models per tier typically providing optimal accuracy-cost tradeoffs.

These ablations collectively demonstrate that \coe's core design principles are empirically sound, with performance gracefully degrading under suboptimal deployment conditions while maintaining substantial benefits when key assumptions (e.g., large relative cost disparities) hold. 

\begin{AIbox}{Takeaway \#5: }
\begin{itemize}[leftmargin=0.7em]
    \setlength\itemsep{0em}
    \item  Ablation studies confirm that \coe's benefits stem from the synergy of parallelization and cost disparities. Safe deferral rules exist across diverse settings, and threshold estimation requires minimal validation data ($\approx$100 samples) while remaining robust across model accuracies.
\end{itemize}
\end{AIbox}

\section{Conclusion}
\begin{Updated}
In this work, we introduce Agreement-Based Cascading (\coe) as a
straightforward approach for adaptive inference that utilizes existing models for
constructing cascades and makes deferral decisions based on their mutual
agreement. 
We define safe deferral rules, ensuring~\coe can serve as a drop-in
replacement for models while improving accuracy.

Although using an ensemble of models can provide a powerful deferral rule for cascading, the additional costs required to compute such an ensemble may not lead to savings in all inference scenarios. Despite this, our work shows that this simple approach is surprisingly effective given the large differences in model sizes that reach state-of-the-art accuracy in recent ML tasks. We demonstrate improvements via a number of real-world case studies, including a study on communication costs in edge-to-cloud inference, rental costs in cloud-based settings, and the cost of black-box API services. 
Overall, our results demonstrate \coe's capacity to improve the efficiency of adaptive
inference systems without the complexities associated with traditional cascade
frameworks, making it a compelling option for practitioners focused on reducing
inference latency. 

\textbf{Future Work.} Several promising directions emerge from this work. First, extending \coe to open-ended generation tasks would significantly broaden its applicability, particularly given the growing prominence of LLM-based applications. 
Second, exploring more efficient ensemble methods could further enhance \coe's benefits in scenarios with limited parallelization capabilities, potentially through techniques that reduce ensemble overhead while maintaining agreement quality. 
Finally, investigating bias-aware agreement mechanisms or construction strategies that promote diversity in decision-making patterns across demographic groups could address potential bias propagation concerns inherent in majority-based voting systems.
\end{Updated}



\section*{Acknowledgements}    
We thank Xiaofang Wang, Wittawat Jitkrittum, Lucio Dery, Pratiksha Thaker, and Kevin Kuo for their helpful comments. This work was supported in part by the National Science Foundation grants IIS2145670, CCF2107024, IIS1705121, IIS1838017, IIS2046613, IIS2112471, and funding from Amazon, Apple, Google, Intel, Meta, Morgan Stanley, Scribe, and the CyLab Security and Privacy Institute. Any opinions, findings and conclusions or recommendations expressed in this material are those of the author(s) and do not necessarily reflect the views of any of these funding agencies.


\clearpage
\bibliography{main_bib}

\begin{thebibliography}{107}
\providecommand{\natexlab}[1]{#1}
\providecommand{\url}[1]{\texttt{#1}}
\expandafter\ifx\csname urlstyle\endcsname\relax
  \providecommand{\doi}[1]{doi: #1}\else
  \providecommand{\doi}{doi: \begingroup \urlstyle{rm}\Url}\fi

\bibitem[Abdalla et~al.(2023)Abdalla, Wahle, Ruas, N{\'e}v{\'e}ol, Ducel, Mohammad, and Fort]{abdalla2023elephant}
Mohamed Abdalla, Jan~Philip Wahle, Terry~Lima Ruas, Aur{\'e}lie N{\'e}v{\'e}ol, Fanny Ducel, Saif~M Mohammad, and Karen Fort.
\newblock The elephant in the room: Analyzing the presence of big tech in natural language processing research.
\newblock In \emph{The 61st Annual Meeting Of The Association For Computational Linguistics}, 2023.

\bibitem[Angelova et~al.(2015)Angelova, Krizhevsky, Vanhoucke, Ogale, and Ferguson]{angelova_real-time_2015}
Anelia Angelova, Alex Krizhevsky, Vincent Vanhoucke, Abhijit~S. Ogale, and Dave Ferguson.
\newblock Real-time pedestrian detection with deep network cascades.
\newblock volume~2, pp.\ ~4, 2015.

\bibitem[Bolukbasi et~al.(2017)Bolukbasi, Wang, Dekel, and Saligrama]{bolukbasi2017adaptive}
Tolga Bolukbasi, Joseph Wang, Ofer Dekel, and Venkatesh Saligrama.
\newblock Adaptive neural networks for efficient inference.
\newblock In \emph{International Conference on Machine Learning}, pp.\  527--536. PMLR, 2017.

\bibitem[Cai et~al.(2019)Cai, Gan, Wang, Zhang, and Han]{cai_once-for-all_2020}
Han Cai, Chuang Gan, Tianzhe Wang, Zhekai Zhang, and Song Han.
\newblock Once-for-{All}: {Train} {One} {Network} and {Specialize} it for {Efficient} {Deployment}, April 2019.

\bibitem[Cai et~al.(2015)Cai, Saberian, and Vasconcelos]{cai_learning_2015}
Zhaowei Cai, Mohammad Saberian, and Nuno Vasconcelos.
\newblock Learning {Complexity}-{Aware} {Cascades} for {Deep} {Pedestrian} {Detection}.
\newblock pp.\  3361--3369, 2015.

\bibitem[Chen et~al.(2020)Chen, Zaharia, and Zou]{chen_frugalml_2020}
Lingjiao Chen, Matei Zaharia, and James~Y Zou.
\newblock {FrugalML}: {How} to use {ML} {Prediction} {APIs} more accurately and cheaply.
\newblock In \emph{Advances in {Neural} {Information} {Processing} {Systems}}, 2020.

\bibitem[Chen et~al.(2023)Chen, Zaharia, and Zou]{chen2023frugalgpt}
Lingjiao Chen, Matei Zaharia, and James Zou.
\newblock {FrugalGPT}: How to use large language models while reducing cost and improving performance.
\newblock \emph{arXiv preprint arXiv:2305.05176}, 2023.

\bibitem[Chen et~al.(2021)Chen, Peng, Fu, and Ling]{Chen_2021_ICCV}
Minghao Chen, Houwen Peng, Jianlong Fu, and Haibin Ling.
\newblock Autoformer: Searching transformers for visual recognition.
\newblock In \emph{Proceedings of the IEEE/CVF International Conference on Computer Vision}, pp.\  12270--12280, October 2021.

\bibitem[Clark et~al.(2020)Clark, Luong, Le, and Manning]{clark2020electra}
Kevin Clark, Minh-Thang Luong, Quoc~V Le, and Christopher~D Manning.
\newblock Electra: Pre-training text encoders as discriminators rather than generators.
\newblock \emph{arXiv preprint arXiv:2003.10555}, 2020.

\bibitem[Cobbe et~al.(2021)Cobbe, Kosaraju, Bavarian, Chen, Jun, Kaiser, Plappert, Tworek, Hilton, Nakano, Hesse, and Schulman]{cobbe2021gsm8k}
Karl Cobbe, Vineet Kosaraju, Mohammad Bavarian, Mark Chen, Heewoo Jun, Lukasz Kaiser, Matthias Plappert, Jerry Tworek, Jacob Hilton, Reiichiro Nakano, Christopher Hesse, and John Schulman.
\newblock Training verifiers to solve math word problems.
\newblock \emph{arXiv preprint arXiv:2110.14168}, 2021.

\bibitem[Crago \& Walters(2015)Crago and Walters]{crago2015heterogeneous}
Stephen~P Crago and John~Paul Walters.
\newblock Heterogeneous cloud computing: The way forward.
\newblock \emph{Computer}, 48\penalty0 (01):\penalty0 59--61, 2015.

\bibitem[Deng et~al.(2009)Deng, Dong, Socher, Li, Li, and Fei-Fei]{5206848}
Jia Deng, Wei Dong, Richard Socher, Li-Jia Li, Kai Li, and Li~Fei-Fei.
\newblock Imagenet: A large-scale hierarchical image database.
\newblock In \emph{2009 IEEE Conference on Computer Vision and Pattern Recognition}, pp.\  248--255, 2009.

\bibitem[Dennis et~al.(2023)Dennis, Shetty, Sevekari, Koishida, and Smith]{dennis_progressive_2023}
Don Dennis, Abhishek Shetty, Anish~Prasad Sevekari, Kazuhito Koishida, and Virginia Smith.
\newblock Progressive ensemble distillation: Building ensembles for efficient inference.
\newblock \emph{Advances in Neural Information Processing Systems}, 36, 2023.

\bibitem[Devlin et~al.(2019)Devlin, Chang, Lee, and Toutanova]{Devlin2019BERTPO}
Jacob Devlin, Ming-Wei Chang, Kenton Lee, and Kristina Toutanova.
\newblock Bert: Pre-training of deep bidirectional transformers for language understanding.
\newblock In \emph{North American Chapter of the Association for Computational Linguistics}, 2019.

\bibitem[Devvrit et~al.(2023)Devvrit, Kudugunta, Kusupati, Dettmers, Chen, Dhillon, Tsvetkov, Hajishirzi, Kakade, Farhadi, and Jain]{devvrit_matformer_2023}
Devvrit, Sneha Kudugunta, Aditya Kusupati, Tim Dettmers, Kaifeng Chen, Inderjit Dhillon, Yulia Tsvetkov, Hannaneh Hajishirzi, Sham Kakade, Ali Farhadi, and Prateek Jain.
\newblock {MatFormer}: {Nested} {Transformer} for {Elastic} {Inference}, October 2023.

\bibitem[Dietterich(2000)]{dietterich2000ensemble}
Thomas~G Dietterich.
\newblock Ensemble methods in machine learning.
\newblock In \emph{International workshop on multiple classifier systems}, pp.\  1--15. Springer, 2000.

\bibitem[Ding et~al.(2024)Ding, Mallick, Wang, Sim, Mukherjee, R{\"u}hle, Lakshmanan, and Awadallah]{dinghybrid}
Dujian Ding, Ankur Mallick, Chi Wang, Robert Sim, Subhabrata Mukherjee, Victor R{\"u}hle, Laks~VS Lakshmanan, and Ahmed~Hassan Awadallah.
\newblock Hybrid llm: Cost-efficient and quality-aware query routing.
\newblock In \emph{The Twelfth International Conference on Learning Representations}, 2024.

\bibitem[Dosovitskiy(2020)]{dosovitskiy2020image}
Alexey Dosovitskiy.
\newblock An image is worth 16x16 words: Transformers for image recognition at scale.
\newblock \emph{arXiv preprint arXiv:2010.11929}, 2020.

\bibitem[Du \& Kaelbling(2024)Du and Kaelbling]{du_compositional_2024}
Yilun Du and Leslie Kaelbling.
\newblock Compositional {Generative} {Modeling}: {A} {Single} {Model} is {Not} {All} {You} {Need}, February 2024.
\newblock arXiv:2402.01103 [cs].

\bibitem[Dubey et~al.(2024)Dubey, Jauhri, Pandey, Kadian, Al-Dahle, Letman, Mathur, Schelten, Yang, Fan, et~al.]{dubey2024llama}
Abhimanyu Dubey, Abhinav Jauhri, Abhinav Pandey, Abhishek Kadian, Ahmad Al-Dahle, Aiesha Letman, Akhil Mathur, Alan Schelten, Amy Yang, Angela Fan, et~al.
\newblock The llama 3 herd of models.
\newblock \emph{arXiv preprint arXiv:2407.21783}, 2024.

\bibitem[D{\v{z}}eroski \& {\v{Z}}enko(2004)D{\v{z}}eroski and {\v{Z}}enko]{dvzeroski2004combining}
Saso D{\v{z}}eroski and Bernard {\v{Z}}enko.
\newblock Is combining classifiers with stacking better than selecting the best one?
\newblock \emph{Machine learning}, 54:\penalty0 255--273, 2004.

\bibitem[Enomoro \& Eda(2021)Enomoro and Eda]{enomoro2021learning}
Shohei Enomoro and Takeharu Eda.
\newblock Learning to cascade: Confidence calibration for improving the accuracy and computational cost of cascade inference systems.
\newblock In \emph{Proceedings of the AAAI Conference on Artificial Intelligence}, volume~35, pp.\  7331--7339, 2021.

\bibitem[Fern \& Givan(2003)Fern and Givan]{fern2003online}
Alan Fern and Robert Givan.
\newblock Online ensemble learning: An empirical study.
\newblock \emph{Machine Learning}, 53:\penalty0 71--109, 2003.

\bibitem[Forooghifar et~al.(2019)Forooghifar, Aminifar, and Atienza]{forooghifar2019resource}
Farnaz Forooghifar, Amir Aminifar, and David Atienza.
\newblock Resource-aware distributed epilepsy monitoring using self-awareness from edge to cloud.
\newblock \emph{IEEE transactions on biomedical circuits and systems}, 2019.

\bibitem[Gangrade et~al.(2021)Gangrade, Kag, and Saligrama]{gangrade_selective_2021}
Aditya Gangrade, Anil Kag, and Venkatesh Saligrama.
\newblock Selective {Classification} via {One}-{Sided} {Prediction}.
\newblock In \emph{Proceedings of {The} 24th {International} {Conference} on {Artificial} {Intelligence} and {Statistics}}, pp.\  2179--2187. PMLR, March 2021.
\newblock ISSN: 2640-3498.

\bibitem[Geifman \& El-Yaniv(2019)Geifman and El-Yaniv]{geifman_selectivenet_2019}
Yonatan Geifman and Ran El-Yaniv.
\newblock {SelectiveNet}: {A} {Deep} {Neural} {Network} with an {Integrated} {Reject} {Option}.
\newblock In \emph{Proceedings of the 36th {International} {Conference} on {Machine} {Learning}}, pp.\  2151--2159. PMLR, May 2019.
\newblock ISSN: 2640-3498.

\bibitem[Geng et~al.(2021)Geng, Gao, Fu, and Zhang]{geng2021romebert}
Shijie Geng, Peng Gao, Zuohui Fu, and Yongfeng Zhang.
\newblock Romebert: Robust training of multi-exit bert.
\newblock \emph{arXiv preprint arXiv:2101.09755}, 2021.

\bibitem[Gontijo-Lopes et~al.(2022)Gontijo-Lopes, Dauphin, and Cubuk]{gontijo-lopes_no_2022}
Raphael Gontijo-Lopes, Yann Dauphin, and Ekin~D. Cubuk.
\newblock No {One} {Representation} to {Rule} {Them} {All}: {Overlapping} {Features} of {Training} {Methods}, April 2022.
\newblock arXiv:2110.12899 [cs].

\bibitem[Goyal et~al.(2020)Goyal, Raghunathan, Jain, Simhadri, and Jain]{goyal2020drocc}
Sachin Goyal, Aditi Raghunathan, Moksh Jain, Harsha~Vardhan Simhadri, and Prateek Jain.
\newblock Drocc: Deep robust one-class classification.
\newblock In \emph{International conference on machine learning}. PMLR, 2020.

\bibitem[Guan et~al.(2018)Guan, Liu, Liu, and Peng]{guan_energy-efficient_2018}
Jiaqi Guan, Yang Liu, Qiang Liu, and Jian Peng.
\newblock Energy-efficient {Amortized} {Inference} with {Cascaded} {Deep} {Classifiers}.
\newblock In \emph{Proceedings of the {Twenty}-{Seventh} {International} {Joint} {Conference} on {Artificial} {Intelligence}}, pp.\  2184--2190, Stockholm, Sweden, July 2018. International Joint Conferences on Artificial Intelligence Organization.
\newblock ISBN 978-0-9992411-2-7.

\bibitem[Guo et~al.(2017)Guo, Pleiss, Sun, and Weinberger]{pmlr-v70-guo17a}
Chuan Guo, Geoff Pleiss, Yu~Sun, and Kilian~Q. Weinberger.
\newblock On calibration of modern neural networks.
\newblock In Doina Precup and Yee~Whye Teh (eds.), \emph{Proceedings of the 34th International Conference on Machine Learning}, volume~70 of \emph{Proceedings of Machine Learning Research}, pp.\  1321--1330. PMLR, 06--11 Aug 2017.

\bibitem[Gupta et~al.(2024)Gupta, Narasimhan, Jitkrittum, Rawat, Menon, and Kumar]{gupta2024language}
Neha Gupta, Harikrishna Narasimhan, Wittawat Jitkrittum, Ankit~Singh Rawat, Aditya~Krishna Menon, and Sanjiv Kumar.
\newblock Language model cascades: Token-level uncertainty and beyond.
\newblock In \emph{International Conference on Learning Representations}, 2024.

\bibitem[Hadi et~al.(2024)Hadi, Al~Tashi, Shah, Qureshi, Muneer, Irfan, Zafar, Shaikh, Akhtar, Wu, et~al.]{hadi2024large}
Muhammad~Usman Hadi, Qasem Al~Tashi, Abbas Shah, Rizwan Qureshi, Amgad Muneer, Muhammad Irfan, Anas Zafar, Muhammad~Bilal Shaikh, Naveed Akhtar, Jia Wu, et~al.
\newblock Large language models: a comprehensive survey of its applications, challenges, limitations, and future prospects.
\newblock \emph{Authorea Preprints}, 2024.

\bibitem[Han et~al.(2022)Han, Huang, Song, Yang, Wang, and Wang]{han_dynamic_2022}
Yizeng Han, Gao Huang, Shiji Song, Le~Yang, Honghui Wang, and Yulin Wang.
\newblock Dynamic {Neural} {Networks}: {A} {Survey}.
\newblock \emph{IEEE Transactions on Pattern Analysis and Machine Intelligence}, 44\penalty0 (11):\penalty0 7436--7456, November 2022.
\newblock ISSN 1939-3539.

\bibitem[He et~al.(2016)He, Zhang, Ren, and Sun]{he2016deep}
Kaiming He, Xiangyu Zhang, Shaoqing Ren, and Jian Sun.
\newblock Deep residual learning for image recognition.
\newblock In \emph{Proceedings of the IEEE conference on computer vision and pattern recognition}, pp.\  770--778, 2016.

\bibitem[He et~al.(2020)He, Liu, Gao, and Chen]{he2020deberta}
Pengcheng He, Xiaodong Liu, Jianfeng Gao, and Weizhu Chen.
\newblock Deberta: Decoding-enhanced bert with disentangled attention.
\newblock \emph{arXiv preprint arXiv:2006.03654}, 2020.

\bibitem[Henighan et~al.(2020)Henighan, Kaplan, Katz, Chen, Hesse, Jackson, Jun, Brown, Dhariwal, Gray, et~al.]{henighan2020scaling}
Tom Henighan, Jared Kaplan, Mor Katz, Mark Chen, Christopher Hesse, Jacob Jackson, Heewoo Jun, Tom~B Brown, Prafulla Dhariwal, Scott Gray, et~al.
\newblock Scaling laws for autoregressive generative modeling.
\newblock \emph{arXiv preprint arXiv:2010.14701}, 2020.

\bibitem[Hestness et~al.(2017)Hestness, Narang, Ardalani, Diamos, Jun, Kianinejad, Patwary, Yang, and Zhou]{hestness2017deep}
Joel Hestness, Sharan Narang, Newsha Ardalani, Gregory Diamos, Heewoo Jun, Hassan Kianinejad, Md~Mostofa~Ali Patwary, Yang Yang, and Yanqi Zhou.
\newblock Deep learning scaling is predictable, empirically.
\newblock \emph{arXiv preprint arXiv:1712.00409}, 2017.

\bibitem[Hou et~al.(2020)Hou, Huang, Shang, Jiang, Chen, and Liu]{hou_dynabert_2020}
Lu~Hou, Zhiqi Huang, Lifeng Shang, Xin Jiang, Xiao Chen, and Qun Liu.
\newblock {DynaBERT}: {Dynamic} {BERT} with {Adaptive} {Width} and {Depth}.
\newblock In \emph{Advances in {Neural} {Information} {Processing} {Systems}}, volume~33, pp.\  9782--9793. Curran Associates, Inc., 2020.

\bibitem[Hu et~al.(2020)Hu, Chen, Wang, and Wang]{hutriple}
Ting-Kuei Hu, Tianlong Chen, Haotao Wang, and Zhangyang Wang.
\newblock Triple wins: Boosting accuracy, robustness and efficiency together by enabling input-adaptive inference.
\newblock In \emph{International Conference on Learning Representations}, 2020.

\bibitem[Huang et~al.(2018)Huang, Chen, Li, Wu, van~der Maaten, and Weinberger]{huang_multi-scale_2018}
Gao Huang, Danlu Chen, Tianhong Li, Felix Wu, Laurens van~der Maaten, and Kilian~Q. Weinberger.
\newblock Multi-{Scale} {Dense} {Networks} for {Resource} {Efficient} {Image} {Classification}, June 2018.
\newblock arXiv:1703.09844.

\bibitem[Jetty et~al.(2021)Jetty, Sawhney, and Zamarin]{aws_batch_2021}
Ramesh Jetty, Indy Sawhney, and Simon Zamarin.
\newblock Batch {Inference} at {Scale} with {Amazon} {SageMaker} {\textbar} {AWS} {Architecture} {Blog}, November 2021.
\newblock URL \url{https://aws.amazon.com/blogs/architecture/batch-inference-at-scale-with-amazon-sagemaker/}.

\bibitem[Jiang et~al.(2023)Jiang, Ren, and Lin]{jiang_llm-blender_2023}
Dongfu Jiang, Xiang Ren, and Bill~Yuchen Lin.
\newblock {LLM}-{Blender}: {Ensembling} {Large} {Language} {Models} with {Pairwise} {Ranking} and {Generative} {Fusion}, June 2023.

\bibitem[Jitkrittum et~al.(2023)Jitkrittum, Gupta, Menon, Narasimhan, Rawat, and Kumar]{jitkrittum_when_2023}
Wittawat Jitkrittum, Neha Gupta, Aditya~K. Menon, Harikrishna Narasimhan, Ankit Rawat, and Sanjiv Kumar.
\newblock When {Does} {Confidence}-{Based} {Cascade} {Deferral} {Suffice}?
\newblock \emph{Advances in Neural Information Processing Systems}, 36:\penalty0 9891--9906, December 2023.

\bibitem[Kaplan et~al.(2020)Kaplan, McCandlish, Henighan, Brown, Chess, Child, Gray, Radford, Wu, and Amodei]{Kaplan2020ScalingLF}
Jared Kaplan, Sam McCandlish, T.~J. Henighan, Tom~B. Brown, Benjamin Chess, Rewon Child, Scott Gray, Alec Radford, Jeff Wu, and Dario Amodei.
\newblock Scaling laws for neural language models.
\newblock \emph{ArXiv}, abs/2001.08361, 2020.

\bibitem[Khare et~al.(2023)Khare, Garg, Kalra, Grandhi, Stoica, and Tumanov]{khare_superserve_2023}
Alind Khare, Dhruv Garg, Sukrit Kalra, Snigdha Grandhi, Ion Stoica, and Alexey Tumanov.
\newblock {SuperServe}: {Fine}-{Grained} {Inference} {Serving} for {Unpredictable} {Workloads}, December 2023.
\newblock arXiv:2312.16733 [cs].

\bibitem[Kim et~al.(2023)Kim, Moon, Tabrizi, Lee, Mahoney, Keutzer, and Gholami]{Kim2023AnLC}
Sehoon Kim, Suhong Moon, Ryan Tabrizi, Nicholas Lee, Michael~W. Mahoney, Kurt Keutzer, and Amir Gholami.
\newblock An llm compiler for parallel function calling.
\newblock \emph{ArXiv}, abs/2312.04511, 2023.

\bibitem[Krizhevsky \& Hinton(2009)Krizhevsky and Hinton]{krizhevsky2009learning}
Alex Krizhevsky and Geoffrey Hinton.
\newblock Learning multiple layers of features from tiny images.
\newblock 2009.

\bibitem[Lai et~al.(2022)Lai, Dai, Singapuram, Liu, Zhu, Madhyastha, and Chowdhury]{lai2022fedscale}
Fan Lai, Yinwei Dai, Sanjay Singapuram, Jiachen Liu, Xiangfeng Zhu, Harsha Madhyastha, and Mosharaf Chowdhury.
\newblock Fedscale: Benchmarking model and system performance of federated learning at scale.
\newblock In \emph{International conference on machine learning}, pp.\  11814--11827. PMLR, 2022.

\bibitem[Lambda(2024)]{lambdacloud}
Lambda.
\newblock {GPU} {Cloud} - {VMs} for {Deep} {Learning} {\textbar} {Lambda}, 2024.
\newblock URL \url{https://lambdalabs.com/service/gpu-cloud}.

\bibitem[Lebovitz et~al.(2023)Lebovitz, Cavigelli, Magno, and Muller]{lebovitz_efficient_2023}
Luzian Lebovitz, Lukas Cavigelli, Michele Magno, and Lorenz~K. Muller.
\newblock Efficient {Inference} {With} {Model} {Cascades}.
\newblock \emph{Transactions on Machine Learning Research}, 2023.

\bibitem[Lee et~al.(2023)Lee, Cheng, and Ostendorf]{lee2023orchestrallm}
Chia-Hsuan Lee, Hao Cheng, and Mari Ostendorf.
\newblock Orchestrallm: Efficient orchestration of language models for dialogue state tracking.
\newblock \emph{arXiv preprint arXiv:2311.09758}, 2023.

\bibitem[Li et~al.(2021{\natexlab{a}})Li, Gadepally, Samsi, Veillette, and Tiwari]{li2021serving}
Baolin Li, Vijay Gadepally, Siddharth Samsi, Mark Veillette, and Devesh Tiwari.
\newblock Serving machine learning inference using heterogeneous hardware.
\newblock In \emph{2021 IEEE High Performance Extreme Computing Conference (HPEC)}, pp.\  1--8. IEEE, 2021{\natexlab{a}}.

\bibitem[Li et~al.(2023)Li, Samsi, Gadepally, and Tiwari]{li2023kairos}
Baolin Li, Siddharth Samsi, Vijay Gadepally, and Devesh Tiwari.
\newblock Kairos: Building cost-efficient machine learning inference systems with heterogeneous cloud resources.
\newblock In \emph{Proceedings of the 32nd International Symposium on High-Performance Parallel and Distributed Computing}, pp.\  3--16, 2023.

\bibitem[Li et~al.(2021{\natexlab{b}})Li, Lin, Chen, Ren, Li, Zhou, and Sun]{li_cascadebert_2021}
Lei Li, Yankai Lin, Deli Chen, Shuhuai Ren, Peng Li, Jie Zhou, and Xu~Sun.
\newblock {CascadeBERT}: {Accelerating} {Inference} of {Pre}-trained {Language} {Models} via {Calibrated} {Complete} {Models} {Cascade}.
\newblock In \emph{Findings of the {Association} for {Computational} {Linguistics}: {EMNLP} 2021}, pp.\  475--486, Punta Cana, Dominican Republic, November 2021{\natexlab{b}}. Association for Computational Linguistics.
\newblock \doi{10.18653/v1/2021.findings-emnlp.43}.

\bibitem[Liu et~al.(2020)Liu, Zhou, Zhao, Wang, Deng, and Ju]{liu2020fastbert}
Weijie Liu, Peng Zhou, Zhe Zhao, Zhiruo Wang, Haotang Deng, and Qi~Ju.
\newblock Fastbert: a self-distilling bert with adaptive inference time.
\newblock \emph{arXiv preprint arXiv:2004.02178}, 2020.

\bibitem[Liu et~al.(2019)Liu, Ott, Goyal, Du, Joshi, Chen, Levy, Lewis, Zettlemoyer, and Stoyanov]{liu2019roberta}
Yinhan Liu, Myle Ott, Naman Goyal, Jingfei Du, Mandar Joshi, Danqi Chen, Omer Levy, Mike Lewis, Luke Zettlemoyer, and Veselin Stoyanov.
\newblock Roberta: A robustly optimized bert pretraining approach.
\newblock \emph{arXiv preprint arXiv:1907.11692}, 2019.

\bibitem[Madaan et~al.(2023)Madaan, Aggarwal, Anand, Potharaju, Mishra, Zhou, Gupta, Rajagopal, Kappaganthu, Yang, Upadhyay, Mausam, and Faruqui]{madaan_automix_2023}
Aman Madaan, Pranjal Aggarwal, Ankit Anand, Srividya~Pranavi Potharaju, Swaroop Mishra, Pei Zhou, Aditya Gupta, Dheeraj Rajagopal, Karthik Kappaganthu, Yiming Yang, Shyam Upadhyay, Mausam, and Manaal Faruqui.
\newblock {AutoMix}: {Automatically} {Mixing} {Language} {Models}, October 2023.
\newblock arXiv:2310.12963 [cs].

\bibitem[Mamou et~al.(2022)Mamou, Pereg, Wasserblat, and Schwartz]{mamou_tangobert_2022}
Jonathan Mamou, Oren Pereg, Moshe Wasserblat, and Roy Schwartz.
\newblock {TangoBERT}: {Reducing} {Inference} {Cost} by using {Cascaded} {Architecture}, April 2022.
\newblock arXiv:2204.06271 [cs].

\bibitem[Miao et~al.(2023)Miao, Oliaro, Zhang, Cheng, Jin, Chen, and Jia]{Miao2023TowardsEG}
Xupeng Miao, Gabriele Oliaro, Zhihao Zhang, Xinhao Cheng, Hongyi Jin, Tianqi Chen, and Zhihao Jia.
\newblock Towards efficient generative large language model serving: A survey from algorithms to systems.
\newblock \emph{ArXiv}, abs/2312.15234, 2023.

\bibitem[Mo et~al.(2023)Mo, Zhu, Shi, Tan, and Wang]{mo2023hetsev}
Hao Mo, Ligu Zhu, Lei Shi, Songfu Tan, and Suping Wang.
\newblock Hetsev: Exploiting heterogeneity-aware autoscaling and resource-efficient scheduling for cost-effective machine-learning model serving.
\newblock \emph{Electronics}, 12\penalty0 (1):\penalty0 240, 2023.

\bibitem[Narasimhan et~al.(2022)Narasimhan, Jitkrittum, Menon, Rawat, and Kumar]{narasimhan_post-hoc_2022}
Harikrishna Narasimhan, Wittawat Jitkrittum, Aditya~K. Menon, Ankit Rawat, and Sanjiv Kumar.
\newblock Post-hoc estimators for learning to defer to an expert.
\newblock \emph{Advances in Neural Information Processing Systems}, 35:\penalty0 29292--29304, December 2022.

\bibitem[Nie et~al.(2024)Nie, Ding, Hu, Jermaine, and Chaudhuri]{nie2024online}
Lunyiu Nie, Zhimin Ding, Erdong Hu, Christopher Jermaine, and Swarat Chaudhuri.
\newblock Online cascade learning for efficient inference over streams.
\newblock \emph{arXiv preprint arXiv:2402.04513}, 2024.

\bibitem[Ong et~al.(2024)Ong, Almahairi, Wu, Chiang, Wu, Gonzalez, Kadous, and Stoica]{ong2024routellm}
Isaac Ong, Amjad Almahairi, Vincent Wu, Wei-Lin Chiang, Tianhao Wu, Joseph~E Gonzalez, M~Waleed Kadous, and Ion Stoica.
\newblock Routellm: Learning to route llms with preference data.
\newblock \emph{arXiv preprint arXiv:2406.18665}, 2024.

\bibitem[Pei et~al.(2021)Pei, Mbakwe, Gupta, Alamir, Lin, Liu, and Shah]{yulong_pei__2021}
Yulong Pei, Amarachi~B. Mbakwe, Abhinav Gupta, Salwa Alamir, Huang Lin, Xiaomo Liu, and Sameena Shah.
\newblock Tweetfinsent: A dataset of stock sentiments on twitter.
\newblock 2021.

\bibitem[Radford et~al.(2021)Radford, Kim, Hallacy, Ramesh, Goh, Agarwal, Sastry, Askell, Mishkin, Clark, et~al.]{radford2021learning}
Alec Radford, Jong~Wook Kim, Chris Hallacy, Aditya Ramesh, Gabriel Goh, Sandhini Agarwal, Girish Sastry, Amanda Askell, Pamela Mishkin, Jack Clark, et~al.
\newblock Learning transferable visual models from natural language supervision.
\newblock In \emph{International conference on machine learning}, pp.\  8748--8763. PMLR, 2021.

\bibitem[Reddy et~al.(2019)Reddy, Chen, and Manning]{reddy2019coqa}
Siva Reddy, Danqi Chen, and Christopher~D Manning.
\newblock Coqa: A conversational question answering challenge.
\newblock \emph{Transactions of the Association for Computational Linguistics}, 7:\penalty0 249--266, 2019.

\bibitem[Rowley et~al.(1998)Rowley, Baluja, and Kanade]{rowley_neural_1998}
H.A. Rowley, S.~Baluja, and T.~Kanade.
\newblock Neural network-based face detection.
\newblock \emph{IEEE Transactions on Pattern Analysis and Machine Intelligence}, 20\penalty0 (1):\penalty0 23--38, January 1998.
\newblock ISSN 1939-3539.
\newblock \doi{10.1109/34.655647}.

\bibitem[{\v{S}}akota et~al.(2024){\v{S}}akota, Peyrard, and West]{vsakota2024fly}
Marija {\v{S}}akota, Maxime Peyrard, and Robert West.
\newblock Fly-swat or cannon? cost-effective language model choice via meta-modeling.
\newblock In \emph{Proceedings of the 17th ACM International Conference on Web Search and Data Mining}, pp.\  606--615, 2024.

\bibitem[Sanh et~al.(2019)Sanh, Debut, Chaumond, and Wolf]{sanh_distilbert_2019}
Victor Sanh, Lysandre Debut, Julien Chaumond, and Thomas Wolf.
\newblock {DistilBERT}, a distilled version of {BERT}: smaller, faster, cheaper and lighter.
\newblock \emph{ArXiv}, October 2019.

\bibitem[Schuster et~al.(2022)Schuster, Fisch, Gupta, Dehghani, Bahri, Tran, Tay, and Metzler]{schuster_confident_2022}
Tal Schuster, Adam Fisch, Jai Gupta, Mostafa Dehghani, Dara Bahri, Vinh Tran, Yi~Tay, and Donald Metzler.
\newblock Confident {Adaptive} {Language} {Modeling}.
\newblock \emph{Advances in Neural Information Processing Systems}, 35:\penalty0 17456--17472, December 2022.

\bibitem[Shafiee et~al.(2018)Shafiee, Shafiee, and Wong]{shafiee2018efficient}
Mohammad~Saeed Shafiee, Mohammad~Javad Shafiee, and Alexander Wong.
\newblock Efficient inference on deep neural networks by dynamic representations and decision gates.
\newblock \emph{Advances in neural information processing systems}, 2018.

\bibitem[Sharkey(1996)]{doi:10.1080/095400996116785}
Amanda J.~C. Sharkey.
\newblock On combining artificial neural nets.
\newblock \emph{Connection Science}, 8\penalty0 (3-4):\penalty0 299--314, 1996.
\newblock \doi{10.1080/095400996116785}.

\bibitem[Shnitzer et~al.(2023)Shnitzer, Ou, Silva, Soule, Sun, Solomon, Thompson, and Yurochkin]{shnitzer_large_2023}
Tal Shnitzer, Anthony Ou, Mírian Silva, Kate Soule, Yuekai Sun, Justin Solomon, Neil Thompson, and Mikhail Yurochkin.
\newblock Large {Language} {Model} {Routing} with {Benchmark} {Datasets}, September 2023.
\newblock arXiv:2309.15789 [cs].

\bibitem[Sinha \& Khandait(2021)Sinha and Khandait]{sinha2021impact}
Ankur Sinha and Tanmay Khandait.
\newblock Impact of news on the commodity market: Dataset and results.
\newblock In \emph{Advances in Information and Communication: Proceedings of the 2021 Future of Information and Communication Conference (FICC), Volume 2}, pp.\  589--601. Springer, 2021.

\bibitem[Socher et~al.(2013)Socher, Perelygin, Wu, Chuang, Manning, Ng, and Potts]{socher2013recursive}
Richard Socher, Alex Perelygin, Jean Wu, Jason Chuang, Christopher~D Manning, Andrew~Y Ng, and Christopher Potts.
\newblock Recursive deep models for semantic compositionality over a sentiment treebank.
\newblock In \emph{Proceedings of the 2013 conference on empirical methods in natural language processing}, pp.\  1631--1642, 2013.

\bibitem[Soo(2014)]{soo_object_2014}
Sander Soo.
\newblock Object detection using {Haar}-cascade {Classifier}.
\newblock \emph{Institute of Computer Science, University of Tartu}, 2\penalty0 (3):\penalty0 1--12, 2014.

\bibitem[Streeter(2018)]{streeter_approximation_2018}
Matthew Streeter.
\newblock Approximation {Algorithms} for {Cascading} {Prediction} {Models}.
\newblock In \emph{Proceedings of the 35th {International} {Conference} on {Machine} {Learning}}, pp.\  4752--4760. PMLR, July 2018.
\newblock ISSN: 2640-3498.

\bibitem[Strubell et~al.(2020)Strubell, Ganesh, and McCallum]{strubell2020energy}
Emma Strubell, Ananya Ganesh, and Andrew McCallum.
\newblock Energy and policy considerations for modern deep learning research.
\newblock In \emph{Proceedings of the AAAI conference on artificial intelligence}, volume~34, pp.\  13693--13696, 2020.

\bibitem[Suggala et~al.(2020)Suggala, Liu, and Ravikumar]{NEURIPS2020_arun}
Arun Suggala, Bingbin Liu, and Pradeep Ravikumar.
\newblock Generalized boosting.
\newblock In H.~Larochelle, M.~Ranzato, R.~Hadsell, M.F. Balcan, and H.~Lin (eds.), \emph{Advances in Neural Information Processing Systems}, volume~33, pp.\  8787--8797, 2020.

\bibitem[Sun et~al.(2022)Sun, Shao, Qian, Huang, and Qiu]{sun2022black}
Tianxiang Sun, Yunfan Shao, Hong Qian, Xuanjing Huang, and Xipeng Qiu.
\newblock Black-box tuning for language-model-as-a-service.
\newblock In \emph{International Conference on Machine Learning}, pp.\  20841--20855. PMLR, 2022.

\bibitem[Susnjak et~al.(2012)Susnjak, Barczak, and Hawick]{Susnjak2012AdaptiveCO}
Teo Susnjak, Andre L.~C. Barczak, and Kenneth~A. Hawick.
\newblock Adaptive cascade of boosted ensembles for face detection in concept drift.
\newblock \emph{Neural Computing and Applications}, 21:\penalty0 671--682, 2012.

\bibitem[Szegedy(2013)]{szegedy2013intriguing}
C~Szegedy.
\newblock Intriguing properties of neural networks.
\newblock \emph{arXiv preprint arXiv:1312.6199}, 2013.

\bibitem[Team et~al.(2024)Team, Riviere, Pathak, Sessa, Hardin, Bhupatiraju, Hussenot, Mesnard, Shahriari, Ram{\'e}, et~al.]{team2024gemma}
Gemma Team, Morgane Riviere, Shreya Pathak, Pier~Giuseppe Sessa, Cassidy Hardin, Surya Bhupatiraju, L{\'e}onard Hussenot, Thomas Mesnard, Bobak Shahriari, Alexandre Ram{\'e}, et~al.
\newblock Gemma 2: Improving open language models at a practical size.
\newblock \emph{arXiv preprint arXiv:2408.00118}, 2024.

\bibitem[Varshney \& Baral(2022)Varshney and Baral]{varshney_model_2022}
Neeraj Varshney and Chitta Baral.
\newblock Model {Cascading}: {Towards} {Jointly} {Improving} {Efficiency} and {Accuracy} of {NLP} {Systems}.
\newblock In \emph{Proceedings of the 2022 {Conference} on {Empirical} {Methods} in {Natural} {Language} {Processing}}, pp.\  11007--11021, Abu Dhabi, United Arab Emirates, 2022. Association for Computational Linguistics.

\bibitem[Viola \& Jones(2001)Viola and Jones]{viola_rapid_2001}
P.~Viola and M.~Jones.
\newblock Rapid object detection using a boosted cascade of simple features.
\newblock In \emph{Proceedings of the 2001 {IEEE} {Computer} {Society} {Conference} on {Computer} {Vision} and {Pattern} {Recognition}. {CVPR} 2001}, volume~1, December 2001.
\newblock ISSN: 1063-6919.

\bibitem[Viola \& Jones(2004)Viola and Jones]{viola_robust_2004}
Paul Viola and Michael~J. Jones.
\newblock Robust {Real}-{Time} {Face} {Detection}.
\newblock \emph{International Journal of Computer Vision}, 57\penalty0 (2):\penalty0 137--154, May 2004.
\newblock ISSN 1573-1405.
\newblock \doi{10.1023/B:VISI.0000013087.49260.fb}.

\bibitem[Wang et~al.(2011)Wang, Lin, and Metzler]{wang_cascade_2011}
Lidan Wang, Jimmy Lin, and Donald Metzler.
\newblock A cascade ranking model for efficient ranked retrieval.
\newblock In \emph{Proceedings of the 34th international {ACM} {SIGIR} conference on {Research} and development in {Information} {Retrieval}}, pp.\  105--114. ACM, 2011.

\bibitem[Wang et~al.(2021)Wang, Kondratyuk, Christiansen, Kitani, Movshovitz-Attias, and Eban]{wang2021wisdom}
Xiaofang Wang, Dan Kondratyuk, Eric Christiansen, Kris~M Kitani, Yair Movshovitz-Attias, and Elad Eban.
\newblock Wisdom of committees: An overlooked approach to faster and more accurate models.
\newblock In \emph{International Conference on Learning Representations}, 2021.

\bibitem[Wang et~al.(2018{\natexlab{a}})Wang, Luo, Crankshaw, Tumanov, Yu, and Gonzalez]{wang_idk_2018}
Xin Wang, Yujia Luo, Daniel Crankshaw, Alexey Tumanov, Fisher Yu, and Joseph~E. Gonzalez.
\newblock {IDK} {Cascades}: {Fast} {Deep} {Learning} by {Learning} not to {Overthink}, June 2018{\natexlab{a}}.
\newblock arXiv:1706.00885 [cs].

\bibitem[Wang et~al.(2018{\natexlab{b}})Wang, Yu, Dou, Darrell, and Gonzalez]{wang_skipnet_2018}
Xin Wang, Fisher Yu, Zi-Yi Dou, Trevor Darrell, and Joseph~E. Gonzalez.
\newblock {SkipNet}: {Learning} {Dynamic} {Routing} in {Convolutional} {Networks}.
\newblock pp.\  409--424, 2018{\natexlab{b}}.

\bibitem[Wang et~al.(2023)Wang, Chen, Tan, and Guo]{wang2023tabi}
Yiding Wang, Kai Chen, Haisheng Tan, and Kun Guo.
\newblock Tabi: An efficient multi-level inference system for large language models.
\newblock In \emph{Proceedings of the Eighteenth European Conference on Computer Systems}, pp.\  233--248, 2023.

\bibitem[Wei et~al.(2022)Wei, Wang, Schuurmans, Bosma, Xia, Chi, Le, Zhou, et~al.]{wei2022chain}
Jason Wei, Xuezhi Wang, Dale Schuurmans, Maarten Bosma, Fei Xia, Ed~Chi, Quoc~V Le, Denny Zhou, et~al.
\newblock Chain-of-thought prompting elicits reasoning in large language models.
\newblock \emph{Advances in neural information processing systems}, 35:\penalty0 24824--24837, 2022.

\bibitem[Wolf et~al.(2019)Wolf, Debut, Sanh, Chaumond, Delangue, Moi, Cistac, Rault, et~al.]{wolf2019huggingface}
Thomas Wolf, Lysandre Debut, Victor Sanh, Julien Chaumond, Clement Delangue, Anthony Moi, Pierric Cistac, Tim Rault, et~al.
\newblock Huggingface's transformers: State-of-the-art natural language processing.
\newblock \emph{arXiv preprint arXiv:1910.03771}, 2019.

\bibitem[Xin et~al.(2020)Xin, Tang, Lee, Yu, and Lin]{xin2020deebert}
Ji~Xin, Raphael Tang, Jaejun Lee, Yaoliang Yu, and Jimmy Lin.
\newblock Deebert: Dynamic early exiting for accelerating bert inference.
\newblock In \emph{Proceedings of the 58th Annual Meeting of the Association for Computational Linguistics}, pp.\  2246--2251, 2020.

\bibitem[Yang et~al.(2024)Yang, Yang, Hui, Zheng, Yu, Zhou, Li, Li, Liu, Huang, et~al.]{yang2024qwen2}
An~Yang, Baosong Yang, Binyuan Hui, Bo~Zheng, Bowen Yu, Chang Zhou, Chengpeng Li, Chengyuan Li, Dayiheng Liu, Fei Huang, et~al.
\newblock Qwen2 technical report.
\newblock \emph{arXiv preprint arXiv:2407.10671}, 2024.

\bibitem[Yang et~al.(2019)Yang, Dai, Yang, Carbonell, Salakhutdinov, and Le]{Yang2019XLNetGA}
Zhilin Yang, Zihang Dai, Yiming Yang, Jaime~G. Carbonell, Ruslan Salakhutdinov, and Quoc~V. Le.
\newblock Xlnet: Generalized autoregressive pretraining for language understanding.
\newblock In \emph{Neural Information Processing Systems}, 2019.

\bibitem[Yu \& Huang(2019)Yu and Huang]{yu_universally_2019}
Jiahui Yu and Thomas~S. Huang.
\newblock Universally {Slimmable} {Networks} and {Improved} {Training} {Techniques}.
\newblock In \emph{Proceedings of the IEEE/CVF International Conference on Computer Vision}, pp.\  1803--1811, 2019.

\bibitem[Yu et~al.(2018)Yu, Yang, Xu, Yang, and Huang]{yu_slimmable_2018}
Jiahui Yu, Linjie Yang, Ning Xu, Jianchao Yang, and Thomas Huang.
\newblock Slimmable {Neural} {Networks}, December 2018.
\newblock arXiv:1812.08928 [cs].

\bibitem[Yue et~al.(2024)Yue, Zhao, Zhang, Du, and Yao]{yue2024large}
Murong Yue, Jie Zhao, Min Zhang, Liang Du, and Ziyu Yao.
\newblock Large language model cascades with mixture of thought representations for cost-efficient reasoning.
\newblock In \emph{The Twelfth International Conference on Learning Representations}, 2024.

\bibitem[Zellers et~al.(2018)Zellers, Bisk, Schwartz, and Choi]{zellers-etal-2018-swag}
Rowan Zellers, Yonatan Bisk, Roy Schwartz, and Yejin Choi.
\newblock {SWAG}: A large-scale adversarial dataset for grounded commonsense inference.
\newblock In \emph{Proceedings of the 2018 Conference on Empirical Methods in Natural Language Processing}, pp.\  93--104, 2018.

\bibitem[Zheng et~al.(2021)Zheng, Guha, Anderson, Henderson, and Ho]{zheng2021does}
Lucia Zheng, Neel Guha, Brandon~R Anderson, Peter Henderson, and Daniel~E Ho.
\newblock When does pretraining help? assessing self-supervised learning for law and the casehold dataset of 53,000+ legal holdings.
\newblock In \emph{Proceedings of the eighteenth international conference on artificial intelligence and law}, pp.\  159--168, 2021.

\bibitem[Zhou et~al.(2020)Zhou, Xu, Ge, McAuley, Xu, and Wei]{zhou2020bert}
Wangchunshu Zhou, Canwen Xu, Tao Ge, Julian McAuley, Ke~Xu, and Furu Wei.
\newblock Bert loses patience: Fast and robust inference with early exit.
\newblock \emph{Advances in Neural Information Processing Systems}, 33:\penalty0 18330--18341, 2020.

\bibitem[Zhu et~al.(2021)Zhu, Lin, Lu, Lin, and Han]{zhu2021delayed}
Ligeng Zhu, Hongzhou Lin, Yao Lu, Yujun Lin, and Song Han.
\newblock Delayed gradient averaging: Tolerate the communication latency for federated learning.
\newblock \emph{Advances in Neural Information Processing Systems}, 34:\penalty0 29995--30007, 2021.

\bibitem[Zhu et~al.(2023)Zhu, Li, Liu, Ma, and Wang]{zhu2023survey}
Xunyu Zhu, Jian Li, Yong Liu, Can Ma, and Weiping Wang.
\newblock A survey on model compression for large language models.
\newblock \emph{arXiv preprint arXiv:2308.07633}, 2023.

\bibitem[Zuo \& de~With(2005)Zuo and de~With]{10.1007/11558484_4}
Fei Zuo and Peter H.~N. de~With.
\newblock Fast face detection using a cascade of neural network ensembles.
\newblock In \emph{Advanced Concepts for Intelligent Vision Systems Conference}, 2005.

\bibitem[Zuo \& de~With(2008)Zuo and de~With]{Zuo_cascaded}
Fei Zuo and Peter H.~N. de~With.
\newblock Cascaded face detection using neural network ensembles.
\newblock \emph{EURASIP Journal on Advances in Signal Processing}, 2008:\penalty0 1--13, 2008.

\end{thebibliography}
\bibliographystyle{tmlr}


\appendix

\section{Agreement-Based Cascading (\coe) can Improve Accuracy}
\label{app:admissible}

Our construction of \coe, as explained in Section~\ref{sec:abchighlevel} and
\ref{sec:formal} is based on \emph{safe deferral rules}
(Definition~\ref{def:safe}) --- rules such that when a data point is selected at
a lower cascade tier, the inference is accurate with high probability.
This means that if the lower tier of the cascade has a higher accuracy compared
to the largest model, on the `easy' data it predicts on, the overall accuracy of
\coe can increase. For `hard' samples \coe and the largest, SoTA model produces
identical predictions as \coe is internally deferring to this model. We
formalize this notion of accuracy improvement in this section.

Consider a classification problem in the statistical learning framework, where
$\mcX$ is our instance space and $\mcY$ is the label space. Let $r$ denote any
deterministic deferral rule $r : \mcX \rightarrow \set{0, 1}$. Let $\set{H_1,
h_2}$ be two classifiers $h_i: \mcX \rightarrow \mcY$ for $i \in \set{1, 2}$. We
will think of $h_2$ as being the more expensive (better accuracy) model. Consider a distribution $P$ over $\mcX \times \mcY$. The risk of classifier $h_2$ is defined as 
$$
    \risk(h_2) = P(h_2(x) \ne  y)
$$

Let us define a cascaded classifier $\mcM_r : \mcX \rightarrow \mcY$ such that,
\begin{align} \label{eq:1}
\mcM_r(x) = \begin{cases}
    H_1(x) & r(x) = 0\\
    h_2(x) & r(x) = 1.
\end{cases}
\end{align}

Then the risk of the cascaded classifier using rule $r$ is,
\begin{align*}
    \risk(\mcM_r) 
        &= P(\mcM_r(x) \ne  y)
\end{align*}
We wish to understand when we can replace an existing large, expensive
classifier $h_2$ with a cascade such that there is no drop in accuracy. This is useful when we
are extremely sensitive towards accuracy drop and wish to only select models at
the lower-level when we are certain about the classification.  Let us define the
excess risk of using a cascade in-place of the larger model $h_2$ as,
$$
    \erisk(\mcM_r, h_2) = \risk(\mcM_r) - \risk(h_2).
$$
Expanding further, we can express the excess risk as,
\begin{align*}
    \erisk(\mcM_r, h_2) 
        &= P(\mcM_r(x) \ne  y) - P(h_2(x) \ne y)\\
        &= P(\mcM_r(x) \ne y \mid r(x) = 0)P(r(x) = 0)  + P(\mcM_r(x) \ne y \mid r(x) = 1)P(r(x) = 1)
        \\ &\quad -
         P(h_2(x) \ne y)\\
        &= P(H_1(x) \ne y \mid r(x) = 0)P(r(x) = 0)  + P(h_2(x) \ne y \mid r(x) = 1)P(r(x) = 1)
        \\ &\quad -
         P(h_2(x) \ne y \mid r(x) = 0)P(r(x) = 0)  + P(h_2(x) \ne y \mid r(x) = 1)P(r(x) = 1)\\
        &= \Big(P(H_1(x) \ne y \mid r(x) = 0) - P(h_2(x) \ne y \mid r(x) = 0)\Big)P(r(x) = 0)\label{eq2}\numberthis
\end{align*}

As can be seen, the general excess risk here depends on both models. Since our
focus is on accuracy,  any rule that leads to a risk that is no worse than that
of the classifier $h_2$ can be used, and we term these as
\textit{admissible} cascades.

\begin{definition}[Admissible cascades] $M_r(x)$ defined above is an admissible
cascade w.r.t the population distribution $P$ and the reference classifier $h_2$
if, 
$$\erisk(M_r, h_2) \le 0,$$
or equivalently
$$
    P(H_1(x) \ne y \mid r(x) = 0) \le P(h_2(x) \ne y \mid r(x) = 0).
$$
\end{definition}

As mentioned in Section~\ref{sec:formal}, for this work, we specialize to the
case where whenever the smaller model is selected, it is always correct. That
is, $P(H_1(x) \ne y \mid r(x) = 0) = 0$. Of course, such a pair of
smaller-classifier $H_1$ and deferral rule $r$ need not exists. However,
whenever they do, a cascade constructed using \emph{any} $h_2$ is admissible
since, 
$$\forall h_2, R(M_r, h_2) = -P(h_2(x) \ne y \mid r(x) = 0)P(r(x) = 0) \le 0.$$
This implies a \emph{universal nature} of such $(H_1, r)$ for such two tier
cascades as the excess risk of the cascade does not depend on the larger
classifier $h_2$.

Finally, if the ensemble $H_1(x)$ outperforms the SoTA model on the `easy'
samples by some strictly positive $\xi > 0$, such that
$$
    P(H_1(x) \ne y \mid r(x) = 0) \le P(h_2(x) \ne y \mid r(x) = 0) - \xi
$$
Then, from Equation~\ref{eq2}
\begin{align*}
    \erisk(\mcM_r, h_2) 
        &= 
        \Big(P(H_1(x) \ne y \mid r(x) = 0) - P(h_2(x) \ne y \mid r(x) =
        0)\Big)P(r(x) = 0)\\
        &= -\xi P(r(x) = 0) < 0,
\end{align*}
wherever the selection rate satisfies $P(r(x) = 0) \ge 0$. Negative excess risk implies an improvement over the SoTA classifier we were using with the overall amount of
improvement depending on $P(r(x) = 0)$.

\subsection*{Proof of Proposition~\ref{prop:costacc}}
\label{app:proof_prop}

\textbf{Part 1:} By Definition~\ref{def:safe}, we have $P(r(x) = 0, H_1^k(x) \neq y) \leq \epsilon$. The risk of the cascade decomposes as:
\begin{align}
R(\mathcal{M}_r) &= P(r(x) = 0, H_1^k(x) \neq y) + P(r(x) = 1, h_2(x) \neq y)\\
&\leq \epsilon + P(r(x) = 1)R(h_2) \leq R(h_2) + \epsilon
\end{align}

\textbf{Part 2:} The expected cost follows from Equation~1 and conditioning on the deferral decision:
$$E[C(\mathcal{M}_r)] = P(r(x) = 0)C(H_1^k) + P(r(x) = 1)C(h_2) = (k^\rho\gamma + P(r(x) = 1))C(h_2)$$

    

\section{Estimating Agreement Threshold $\theta$}
\label{app:samplecomplexity}


\begin{figure}[h]
\vspace{-5mm}
      \begin{subfigure}[b]{1.0\textwidth}
      \centering
        \includegraphics[width=1.0\textwidth]{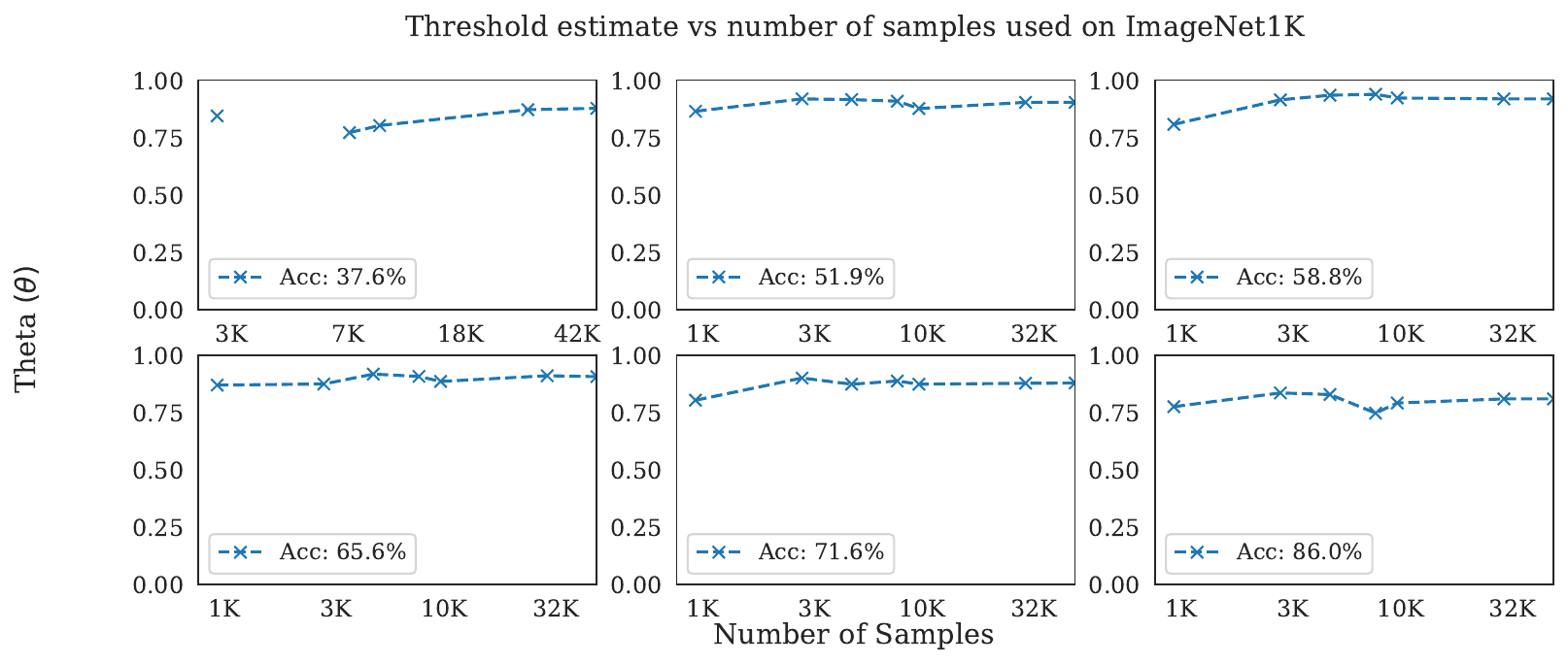}
      \end{subfigure}
\vspace{-4mm}
    \caption{Estimation of agreement threshold $\theta$ stability as a function of the number of samples used, across different model accuracy levels on the ImageNet-1K dataset. Each plot corresponds to a model with a specific accuracy, shown in the legend. The initial estimate is with $100$ samples, with subsequent estimates with larger and larger number of samples. The initial estimate is reasonably close within subsequent estimates with larger number of samples.
    }
    \label{fig:estimationrate}
\vspace{-3mm}
\end{figure}

To effectively apply \coe, it’s essential to configure an agreement threshold
$\theta$ for the deferral rules defined in Equations~\ref{eq:voterules1}
and~\ref{eq:voterules2}. 
This threshold indicates a sufficient level of confidence in the ensemble’s
predictions, allowing \coe to avoid deferring to a higher-cost model when the
ensemble’s agreement is high, thereby reducing inference costs without
sacrificing accuracy. Recall from Definition~\ref{def:safe} that the failure rate of a deferral rule,
as a function of the agreement threshold is given by,
$$
    \prob(s(x) \ge \theta, H_k(x) \ne y).
$$
In particular, safe deferral rules are those with a failure rate bounded by a
small $\eps \ge 0$ of our choice. 
This means that, given
a distribution $\prob$ over $\mcX\times \mcY$ we can define a function $p(\theta)$ as,
$$
    p(\theta) = \prob(s(x) \ge \theta, H_k(x) \ne y).
$$
We can now define feasible thresholds, $\theta$  as those for which the error
rate $p(\theta) \le \eps $, since any such $\theta$ leads to safe deferral.
In practice, we rarely have access to $\prob$ and therefore $p(\theta)$. We
instead use its plugin-estimator, given by
$$
    \hat{p}(\theta) = \frac{1}{n}\sum_{i=1}^{n} \eye[s(x) \ge \theta, H_k(x) \ne y].
$$

We estimate $\hat{p}(\theta)$ by using a small subset of samples from the
validation set. Typically, we draw around 100 samples and set them aside to
determine a stable threshold.  Figure~\ref{fig:estimationrate} shows how the estimated threshold varies with
the number of samples used in the estimation process, across models with
different accuracy levels. Each plot represents a model of a specific accuracy,
with accuracy percentages indicated in the legend. As shown, the estimate generally stabilizes even with 100 samples, suggesting
that only a few validation samples are needed to reliably estimate
$\theta$, making the parameter estimation process efficient and practical for
real-world applications.



    
    

\section{Existence of Safe Deferral Rules and Selection Rates}
\label{app:selection}

\begin{figure}[h]
\vspace{-4mm}
      \begin{subfigure}[b]{0.5\textwidth}
      \centering
        \includegraphics[width=1.0\textwidth]{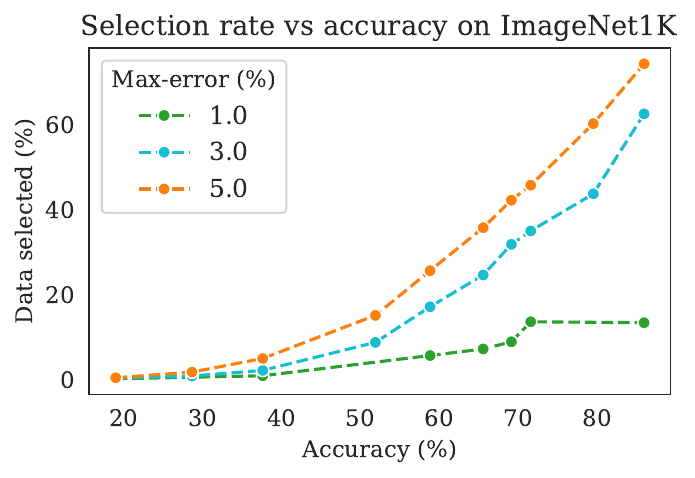}
      \end{subfigure}
      \begin{subfigure}[b]{0.5\textwidth}
      \centering
        \includegraphics[width=1.0\textwidth]{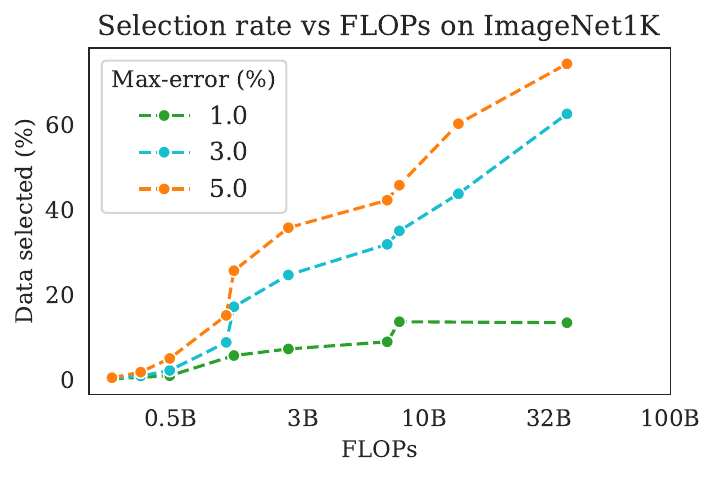}
       \end{subfigure}
\vspace{-3mm}
    \caption{Selection rate as a function of accuracy (left) and FLOPs (right)
    for different error tolerances on ImageNet-1K. The selection rate
    $\prob(r(x)\ge \theta)$ represents the fraction of data handled at a lower
    cascade tier without deferring to a larger model, based on a threshold
    $\theta$. Laxer error tolerances (e.g., 5\%) yield higher selection rates,
    as more samples meet the criteria for safe deferral. In contrast, stricter
    tolerances (e.g., 1\%) result in lower selection rates. Both plots
    illustrate that higher-accuracy or higher-FLOP models generally achieve
    higher selection rates, especially as the allowable error tolerance
    increases, reinforcing the stability and practicality of~\coe.
    }
\vspace{-3mm}
    \label{fig:selection_rate}
\end{figure}

This section examines the existence of safe deferral rules --- rules that have a
very low probability of being incorrect, where data is not deferred to a larger
model --- by evaluating selection rates. For any threshold $\theta$, the
selection rate $P(r(x) \ge \theta)$ represents the fraction of data that deemed
`easy' enough to be processed at a lower tier of the cascade. Experimentally, we
observe that $\theta_\eps$ computed in the manner described in
Section~\ref{app:samplecomplexity} does indeed adhere to the error tolerance of choice $\eps$. This
means that whenever a data point is selected at a lower tier, the inference is
accurate with probability $1-\eps$. Stricter values of $\eps$ typically result
in a higher threshold for agreement and a lower-selection rate.


The plots in Figure~\ref{fig:selection_rate} show the selection rate across
different accuracy levels and FLOP values on the ImageNet-1K dataset. We explore
selection rates for three error tolerances, corresponding to maximum allowable
error rates of 1\%, 3\%, and 5\%. As shown in the left plot, which
compares selection rate versus accuracy of the models used in the ensemble,
higher accuracy models tend to achieve higher selection rates, especially when
the error tolerance is relaxed (e.g., 5\% maximum error). Conversely, stricter
error tolerances (e.g., 1\%) lead to lower selection rates across all accuracy
levels, as fewer samples meet the stringent requirements for safe deferral at
lower cascade tiers. In the right plot, which illustrates selection rate versus
FLOPs, a similar trend is observed. Higher-FLOP models, which are generally more
accurate, are able to safely handle a larger portion of the data at lower tiers,
particularly as the error tolerance increases. 




\section{Evaluation Setups}
\label{app:setup}

\subsection{Datasets and Models}

Datasets and models used in our experiments are detailed in Table~\ref{tab:datasets_used} and Table~\ref{tab:models_used}.
\begin{table}[h]
\vspace{-3mm}
  \centering
  \caption{Datasets used in both benchmark and black-box API experiments across various task types.}
  \begin{tabular}{c c c}
    \toprule
    \textbf{Category} & \textbf{Dataset} & \textbf{Task Type} \\
    \midrule
    \multirow{2}{*}{Image Tasks}  & ImageNet-1K~\citep{5206848} & Image classification \\
    & CIFAR-10~\citep{krizhevsky2009learning} & Image classification \\
                           
    \midrule
    \multirow{3}{*}{Language Tasks} & SST-2~\citep{socher2013recursive} & Sentiment analysis \\
                            & Twitter Financial News~\citep{yulong_pei__2021} & Sentiment analysis \\
                            & SWAG~\citep{zellers-etal-2018-swag} & Multiple-Choice QA \\
    \midrule
    \multirow{4}{*}{Black-Box Experiments}                & GSM8K~\citep{cobbe2021gsm8k} & Math Reasoning \\
    & COQA~\citep{reddy2019coqa} & Conversational QA \\
    & OVERRULING~\citep{zheng2021does} & Legal Reasoning \\
    & HEADLINES~\citep{sinha2021impact} & News Classification\\
    \bottomrule
  \end{tabular}
\vspace{-3mm}
  \label{tab:datasets_used}
\end{table}

\begin{table}[h]
\vspace{-5mm}
  \caption{Summary of models used for both benchmark and black-box API experiments, across image and text tasks.}
\vspace{-3mm}
  \centering
  \begin{tabular}{c c c}
    \toprule
    \textbf{Category} & \textbf{Dataset} & \textbf{Tiers Used} \\
    \midrule
    \multirow{4}{*}{Language Models}  & BERT~\citep{Devlin2019BERTPO} & \\
    & RoBERTa~\citep{liu2019roberta} & \\
    & XLNet~\citep{Yang2019XLNetGA} & \texttt{BASE}, \texttt{LARGE} \\
    & ELECTRA~\citep{radford2021learning} & \\
                           
    \midrule
    \multirow{3}{*}{Image Models} & ResNet~\citep{he2016deep} & \\
    & ViT~\citep{dosovitskiy2020image} & Selection based on FLOPs \\
    & CLIP~\citep{clark2020electra} & \\
    \midrule
    \multirow{3}{*}{Black-Box Models}    & LlaMA~3.1~\citep{dubey2024llama} & \\
    & Gemma~2~\citep{team2024gemma} & See Table \ref{tab:model_costs} \\
    & Qwen~2~\citep{yang2024qwen2} & \\
    \bottomrule
  \end{tabular}
    \vspace{-4mm}
  \label{tab:models_used}
\end{table}

\subsection{Details for Black-Box API Experiments}
\label{app:methodspecific}

\paragraph{Baselines
} We compare \coe to FrugalGPT~\citep{chen2023frugalgpt}, 2
variants of AutoMix~\citep{madaan_automix_2023}, and MoT LLM
Cascade~\citep{yue2024large}. Although HybridLLM~\citep{dinghybrid} falls into
this category of SOTA methods, it has been shown to
underperform FrugalGPT and AutoMix~\citep{madaan_automix_2023}. 
For practical comparison, we implement all methods in a fully functional cascade system.



\paragraph{Method-Specific Details}
\begin{itemize}
    \item \textbf{AutoMix}: AutoMix trains a different router for all possible cascade steps, i.e., $n-1$ routers for the n cascade tiers, and this has to be repeated for every new task or model replacement in the system. A threshold (\textsc{AutoMix+T}) or POMDP (\textsc{AutoMix+P}) is trained with a combination of $\geq50$ training samples from the same test data distribution and the initial inferences generated on the test data by the two models involved in each cascading step. After training the router, using the cascading system often involves running a few-shot self-verification 8 times at a high sampling temperature (temp = 1.0), using the same model that generates inference at the given cascade tier. Automix then averages the self-verification results and meta-verify it with the best parameters of the routing strategy to decide when to exit.
    \item \textbf{FrugalGPT}: Just like AutoMix, FrugalGPT needs to train $n-1$ routers for $n$ cascade tiers, and each router needs to have a sense of the data distribution and the model's predictive power. $\geq500$ training samples and inference generated on these samples by the tier's model are needed to train each tier/model's scorer, a DistilBERT \citep{sanh_distilbert_2019}. 
    \item \textbf{MoT LLM Cascade}: \citet{yue2024large} focuses on sampling and consistency checking as a means of cascade. To measure consistency, the weaker LLM generates multiple answers for a single question by varying the randomness in the LLM's responses---i.e., varying the temperature of the model---while using in-context demonstrations and reasoning techniques (e.g., Chain-of-Thought \citep{wei2022chain}) to influence how the model generates answers. The system compares the different sampled answers and picks the most consistent one. If the consistency score is high enough, the weaker model's answer is accepted. Otherwise, the question is passed to the next tier.
    \item \textbf{\coe}: We use the voting-based safe deferral rule, requiring no additional training or any complex routing strategies.
\end{itemize}

\paragraph{Cascade Models} We use the models described in Table~\ref{tab:model_costs}. We consider a setting that is advantageous to the baselines by selecting the \textit{best singlular model} from each performance tier and cascading between them. In all cascading systems, we use all three tiers for cascade; but considering budget constraints, we also have setups where we delete Tier 3 and use only the first two tiers (2-level cascade).

\paragraph{Evaluation Setup} We evaluate these methods on a variety of (closed) generation datasets and tasks as shown in Table~\ref{tab:datasets_used}. To ensure consistency in output format and easier evaluation setup, we use few-shot prompting—specifically, 4-shot—for all models across all tasks. For evaluation metrics, we use the macro F1 score for \textsc{CoQA} to capture overlaps between predictions and ground-truth answers, while we measure accuracy (essentially exact match) for the rest of the tasks. In terms of efficiency, we also measure the costs of using the model APIs. It is important to note that AutoMix and FrugalGPT incur extra setup costs that we did not factor into our results. These costs and associated latency represent a significant constraint, especially for scenarios requiring frequent retraining or adaptation to new tasks, distributions, and models.

As shown in Figure \ref{fig:api_plots}, we observe that \coe is often more cost-effective than the baselines, even with their sophisticated routing mechanisms and singular model tiers. 
Our analyses show that \coe's efficient deferral strategy allows a more aggressive utilization of cheaper models for a significant portion of the input while only using expensive models when necessary. For instance, we realize---upon analyzing FrugalGPT's results---that the trained scorer struggles as an efficient deferral signal as the tasks get harder; hence, it is more likely to take the safer option to cascade as test sample difficulty increases. This means that \coe can be expected to be more efficient since the scaling laws ensure that the sum of the costs of using several cheaper models is still much less than the cost of using the larger model in the next tier. AutoMix, on the other hand, uses a few-shot self-verification system that is sampled $>1$ times; hence, the additional API calls add significantly to its cost of usage. Considering that the self-verification process is an integral part of the AutoMix setup, it can be guaranteed that \coe will \textit{always} be cheaper to use than AutoMix, despite using more models. 

\section{Benefits of \coe}
\subsection{Parallel vs. sequential inference execution}
\label{app:sequential_parallel}

\begin{figure*}[h]
\vspace{-4mm}
  \begin{subfigure}[b]{0.33\textwidth}
    \centering
    \includegraphics[width=\textwidth]{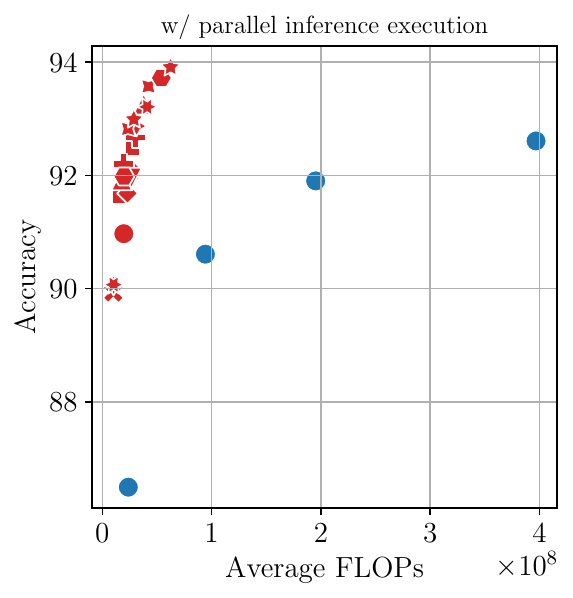}
  \end{subfigure}
  \hfill
  \begin{subfigure}[b]{0.6\textwidth}
    \centering
    \includegraphics[width=\textwidth]{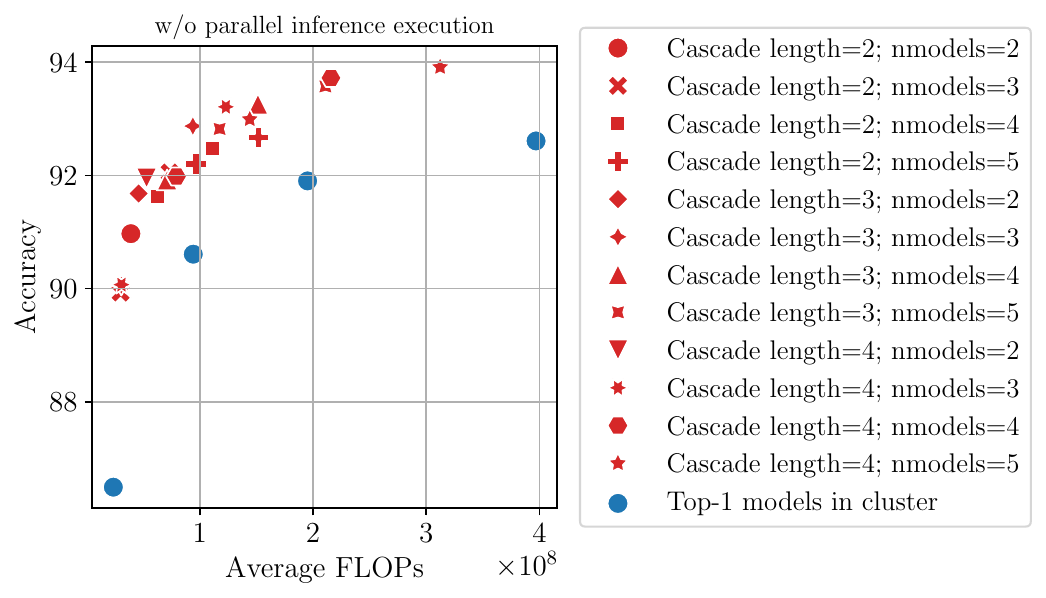}
  \end{subfigure}
  \caption{Impact of parallelization on \coe performance for CIFAR-10. Left: With parallel inference execution ($\rho = 1$), \coe configurations consistently outperform the best single models across different cascade lengths and ensemble sizes. Right: Even with sequential execution ($\rho = 0$), \coe maintains advantages over single models, though with reduced efficiency. The results demonstrate that while parallelization is beneficial, \coe remains effective even under sequential constraints when cost disparities are sufficient.
}
    \vspace{-3mm}
  \label{fig:parallel}
\end{figure*}

Based on Section \ref{sec:freeparallel}, we additionally show in Figure \ref{fig:parallel} the superiority of parallel inference execution for cascading over using the best single model using CIFAR-10 as a case study. We also show that in the worst-case scenarios in which every single inference is sequentially produced over each ensemble and cascade, there are still considerable savings over the largest single models, if the scenarios assumptions are met. 

\subsection{Cost Benefits}
\label{app:cost_analysis}
Based on Section \ref{sec:gpu_costs}, Table \ref{tab:gpu_costs} shows the GPUs' pricing across several tiers retrieved from Lambda Cloud. Table \ref{tab:costs_table} shows a detailed analysis of costs across all cascade tiers, associated with the number of cascade exits at each cascade tier, to provide holistic efficiency comparisons of the aggregated cascade costs against using the best (and largest) model --- and \coe dominates in every measured metric.
Typically, most cascade exits occur in the earlier (and much cheaper) tiers (as also shown in Table \ref{tab:costs_table}), ensuring that the more cost-intensive cascade tiers featuring are reserved for the harder test instances.

\begin{table}[h]
\vspace{-3mm}
  \centering
  \caption{GPU rental costs from Lambda Cloud~\citep{lambdacloud} (September 2024) showing the substantial cost disparities between hardware generations. The 25$\times$ cost difference between H100 and V100 GPUs, combined with more modest throughput differences, creates favorable conditions for \coe's heterogeneous hardware placement strategy described in Section~\ref{sec:gpu_costs}}
  \begin{tabular}{c|c}
    \hline
    \textbf{GPU} & \textbf{Cost per Hour (USD)} \\
    \hline
    V100  & 0.5  \\
    \hline
    A6000 & 0.8  \\
    \hline
    A100  & 1.29 \\
    \hline
    H100  & 2.49 \\
    \hline
  \end{tabular}
\vspace{-4mm}
  \label{tab:gpu_costs}
\end{table}

\begin{table*}[t!]
\centering
\caption{Detailed cost breakdown across cascade tiers for each dataset, showing the fraction of samples processed at each tier, associated GPU costs, latency, and FLOPs. The high fraction of samples processed at cheaper early tiers (52-93\%) demonstrates \coe's effectiveness at concentrating expensive computation on truly difficult samples. \coe consistently outperforms single best models across all metrics while achieving substantial cost savings.}
\label{tab:costs_table}
\resizebox{\linewidth}{!}{%
\renewcommand{\arraystretch}{1.5}
\begin{tabular}{ll|llllll}
\hline
& & & & & & & \textbf{Best Single} \\
\textbf{Dataset} & \textbf{Metric} & \textbf{Tier 1} & \textbf{Tier 2} & \textbf{Tier 3} & \textbf{Tier 4} & \textbf{\coe} & \textbf{Model}  \\ \hline \hline
\textbf{CIFAR-10} & Frac. Samples (total=10,000) & 0.73 & 0.09 & 0.08 & 0.10 & 1.00 & 1.00 \\
                & Total GPU Cost (\$ / hour) & 0.36 & 0.07 & 0.11 & 0.24 & 0.79 & 2.49 \\
                & Avg. Latency (ms) & 3.11 & 3.79 & 7.76 & 9.07 & 4.13 & 9.07 \\
                & Avg. FLOPs & 5.42e6 & 2.32e7 & 1.16e8 & 2.47e8 & 3.97e7  & 2.48e8 \\ \hline
\textbf{ImageNet-1K} & Frac. Samples (total=50,000) & 0.52 & 0.29 & 0.19 & - & 1.00 & 1.00 \\
                & Cost (\$ / hour) & 0.26 & 0.23 & 0.25 & - & 0.74 & 1.29 \\
                & Avg. Latency (ms) & 2.45 & 2.88 & 3.17 & - & 2.71 & 3.17 \\
                & Avg. FLOPs & 2.15e9 & 3.90e9 & 4.30e9 & - & 3.07e9 & 4.30e9 \\ \hline
\textbf{SWAG (MCQ)} & Frac. Samples (total=20,006) & 0.71 & 0.29 & - & - & 1.00 & 1.00 \\
                & Cost (\$ / hour) & 0.36 & 0.23 & - & - & 0.59 & 0.80 \\
                & Avg. Latency (ms) & 4.52 & 8.05 & - & - & 5.53 & 8.05 \\
                & Avg. FLOPs & 1.88e10 & 6.67e10 & - & - & 3.25e10  & 6.67e10 \\ \hline
\textbf{SST-2} & Frac. Samples (total=872) & 0.93 & 0.07 & - & - & 1.00 & 1.00 \\
                & Cost (\$ / hour) & 0.46 & 0.06 & - & - & 0.52 & 0.80 \\
                & Avg. Latency (ms) & 3.88 & 7.22 & - & - & 4.13 & 7.22 \\
                & Avg. FLOPs & 5.43e9 & 1.68e10 & - & - & 6.26e9 & 1.68e10 \\ \hline
\textbf{Twitter Fin News} & Frac. Samples (total=822) & 0.65 & 0.35 & - & - & 1.00 & 1.00 \\
                & Cost (\$ / hour) & 0.32 & 0.28 & - & - & 0.61 & 0.80 \\
                & Avg. Latency (ms) & 4.05 & 7.26 & - & - & 5.19 & 7.26 \\
                & Avg. FLOPs & 6.83e9 & 2.42e10 & - & - & 1.30e10  & 2.42e10 \\ \hline
\end{tabular}%
}
\end{table*}

\end{document}